\documentclass{article}

\usepackage[preprint]{neurips_2026}

\usepackage[utf8]{inputenc}
\usepackage[T1]{fontenc}
\usepackage{hyperref}
\usepackage{url}
\usepackage{booktabs}
\usepackage{array}
\usepackage{graphicx}
\usepackage{amsfonts}
\usepackage{nicefrac}
\usepackage{microtype}
\usepackage{amsmath, amssymb, amsthm}
\usepackage{cleveref}
\usepackage{caption}
\usepackage{mathtools}
\usepackage{multirow}
\usepackage{wrapfig}
\graphicspath{ {figures/} }
\usepackage[nodisplayskipstretch]{setspace}

\theoremstyle{plain}
\newtheorem{theorem}{Theorem}[section]

\theoremstyle{definition}
\newtheorem{definition}[theorem]{Definition}

\theoremstyle{remark}

\newcommand{\ourmethod}{\textsc{ICON}}

\title{Towards Invariance to Node Identifiers in Graph Neural Networks}

\author{
  Maya Bechler-Speicher \\
  Blavatnik School of Computer Science\\
  Tel-Aviv University\\
  \texttt{mayab4@mail.tau.ac.il} \\
  \And
  Moshe Eliasof \\
  Department of Applied Mathematics and Theoretical Physics\\
  University of Cambridge \\
  \And
  Carola-Bibiane Sch\"onlieb \\
  Department of Applied Mathematics and Theoretical Physics\\
  University of Cambridge \\
  \And
  Ran Gilad-Bachrach \\
  Department of Bio-Medical Engineering\\
  Edmond J. Safra Center for Bioinformatics\\
  Tel-Aviv University \\
  \And
  Amir Globerson \\
  Google Research \\
  Blavatnik School of Computer Science\\
  Tel-Aviv University \\
}

\begin{document}

\maketitle

\begin{abstract}
Message-Passing Graph Neural Networks (GNNs) are known to have limited expressive power, due to their message passing structure.
One mechanism for circumventing this limitation is to add unique node identifiers (IDs), which break the symmetries that underlie the expressivity limitation.
In this work, we highlight a key limitation of the ID framework, and propose an approach for addressing it. We begin by observing that the final output of the GNN should clearly not depend on the specific IDs used. We then show that in practice the network does in fact depend on the IDs, which is undesirable. Invariance to node IDs may be enforced in several ways, and we discuss their theoretical properties.
 We then propose a novel regularization method that effectively enforces ID invariance to the network. Extensive evaluations on both real-world and synthetic tasks demonstrate that our approach significantly improves ID invariance and, in turn, often boosts generalization performance.
\end{abstract}

\section{Introduction}
\label{sec:intro}
Graph Neural Networks (GNNs), and in particular Message-Passing Graph Neural Networks (MPGNNs) \citep{morris2021weisfeiler} are limited in their expressive power because they may represent distinct graphs in the same way \citep{pmlr-v119-garg20c}. However, when unique node identifiers (IDs) are provided, this limitation is overcome and the MPGNNs become Turing complete \citep{Loukas2020What, pmlr-v119-garg20c}.

There are several ways to attach unique IDs to nodes. A simple approach is to add features with random values as identifiers and resample them during the training process \citep{sato2021random, ijcai2021p291}. 
This strategy, which we refer to as Random Node Initialization (RNI), almost surely assigns unique IDs to each node.
Consequently, RNIs transform MPGNNs into universal approximators of invariant and equivariant graph functions \citep{ijcai2021p291}.
Nonetheless, RNI has been shown to not improve generalization on real-world datasets \citep{pmlr-v97-murphy19a, you2021identity, gpse_canturk, eliasof2024granola, NEURIPS2021_b8b2926b, bevilacqua2022equivariant, eliasof2023graph}.

Note that although IDs are used as GNN features, specific ID values should not affect network output. Namely, two different assignments to the IDs should ideally result in the same output. Otherwise, different ID assignments would result in different predictions for the graph label, and therefore at least one of them would be wrong. We refer to this desirable property as ``invariance to ID values''. Intuitively, if such an invariance does not hold, generalization is negatively impacted.

Surprisingly, we find that in practice this invariance often does not hold for GNNs learned with RNIs. This is shown across multiple GNN architectures, real-world, and synthetic datasets, even when resampling IDs per batch. 
These results align with prior research on the implicit bias of GNNs trained with gradient-based methods, which demonstrated a tendency to overfit to graph structures even in scenarios where such structures should be disregarded \citep{3692070.3692201}. Consequently, although GNNs are theoretically capable of converging to ID-invariant solutions, they fail to do so.
This lack of invariance leads to overfitting. 
Thus, we set out to improve the ID-invariance of learned GNNs. 

There are several ways of constraining models to be invariant to IDs. For example, one can constrain the output of the first GNN layer to have the same output for all assignments of node IDs. However, we prove that this severely limits the expressive power of the  GNN. Consequently, we conclude that improving expressiveness requires including layers that are not invariant to IDs. Additionally, we demonstrate that a three-layer architecture, with only the final layer being invariant to IDs, suffices to achieve invariance to IDs and expressiveness beyond that of the 1-WL test.

Building on our theoretical findings, we introduce {\ourmethod}: ID-Invariance through Contrast, an approach that achieves ID invariance through explicit regularization. We evaluate \ourmethod~ on both real-world and synthetic datasets across various GNN architectures, demonstrating that \ourmethod~ significantly enhances ID invariance and generalization in many cases. Furthermore, we show that \ourmethod~ improves extrapolation capabilities and accelerates training convergence.

\textbf{Main Contributions:} The goal of this work is to improve the utilization of node IDs in GNNs by making models ID-invariant. Our key contributions are:
\begin{enumerate}
    \item  We show, on multiple real-world datasets, that GNNs with RNI fail to learn invariance to IDs, despite resampling. This is an undesirable property, potentially leading to overfitting.
    
 \item We theoretically analyze the properties of GNNs with IDs and derive concrete and practical requirements to achieve invariance to IDs while improving expressiveness with respect to GNNs without IDs. 
 \item Based on our theoretical analysis, we present \ourmethod - an efficient method, that is compatible with any GNN, to achieve invariance to IDs while improving expressiveness.
 \item  We evaluate {\ourmethod} on real-world and synthetic datasets with a variety of GNNs, and show that it significantly improves invariance to IDs and often improves generalization. Finally, we demonstrate how {\ourmethod} can improve extrapolation capabilities and accelerate training convergence.
\end{enumerate}

\section{Related work}
\label{sec:related}

In this section, we review relevant prior work, focusing on the use of IDs in GNNs, and neural networks with invariance properties.
\subsection{Graph neural networks}
 The key idea of GNNs \citep{kipf2017semisupervised,pmlr-v70-gilmer17a,gat,NIPS2017_5dd9db5e,3737916.3740790} is to use neural networks that combine node features with graph structure to obtain useful graph representations. This combination is done in an iterative manner, which can capture complex properties of the graph and its node features. See \citet{9046288} for a recent survey.

\subsection{Node identifiers in GNNs}
Many GNNs, such as MPGNNs, are inherently limited in their ability to distinguish between nodes with structurally similar neighborhoods. This limitation makes many GNNs only as powerful as the 1-Weisfeiler-Lehman (WL) isomorphism test \citep{morris2021weisfeiler}.

To address this limitation, the assignment of node IDs has emerged as a practical technique to enhance the expressiveness of GNNs \citep{sato2021random, ijcai2021p291, Bouritsas_2023, pellizzoni2024on, gpse_canturk, eliasof2023graph}.
The addition of node IDs can assist the GNN to distinguish between nodes that would otherwise be indistinguishable by message-passing schemes.
For example, \citet{you2021identity} showed that expressiveness improves by adding binary node IDs.
Recently, \citet{pellizzoni2024on} suggested a method for assigning IDs using Tinhofer-based node ordering \citep{TINHOFER1991253, pellizzoni2024on}.
Another approach to add node IDs is to calculate new structural node features, such as counting triangles \citep{Bouritsas_2023} or computing Laplacian eigenvalues \citep{dwivedi2021generalizationtransformernetworksgraphs}, which can sometimes separate nodes that were otherwise indistinguishable.  
These approaches are deterministic, yet do not guarantee the uniqueness of node representation.
To achieve uniqueness almost surely, \citet{sato2021random} and \citet{ijcai2021p291} suggested sampling random node features with sufficient dimension at every epoch. This approach was shown to make MPGNNs universal approximators of invariant and equivariant graph functions \citep{ijcai2021p291}.
In this work we focus on these non-deterministic approaches, and claim that the model should be invariant to the values of IDs. 

\subsection{Invariant neural networks}
The topic of invariance in deep learning has been explored in various domains, including Convolutional Neural Networks (CNNs) and GNNs \citep{Simard1998}. 
In CNNs, translational invariance is achieved by design. However, models such as Spatial Transformer Networks (STN) enable CNNs to handle more complex geometric transformations, such as invariances to rotations and zooming \citep{jaderberg2015spatial}. Furthermore, the Augerino approach, which learns invariances by optimizing over a distribution of data augmentations, is an effective strategy to learn invariances in tasks such as image classification \citep{benton2020learning}. TI-Pooling, introduced in transformation-invariant CNN architectures, further addresses the need for invariance in vision-based tasks \citep{laptev2016tipooling}.
In graph tasks, learning invariances has seen rapid development, particularly with GNNs. Traditionally, GNNs are designed to be permutation invariant, but recent research has also explored explicit mechanisms to learn invariances rather than relying solely on built-in properties. For example, \citet{xia2023learning} introduced a mechanism that explicitly learns invariant representations by transferring node representations across clusters, ensuring generalization under structure shifts in the test data. This method is particularly effective in scenarios where training distributions and test distributions may differ, a problem known as out-of-distribution generalization. 
\citet{jia2024graph}  proposed a strategy to extract invariant patterns by mixing environment-related and invariant subgraphs. 
In this work, we present \ourmethod{} which promotes ID invariance while maintaining their added expressive power.

\section{GNNs overfit IDs}
\label{sec:overfitting}

In this section, we demonstrate that GNNs trained with RNIs fail to converge to solutions that are invariant to the values of the IDs. 
We hypothesize that this lack of invariance may contribute to the poor empirical performance of RNIs in practice, and show in Section~\ref{sec:experiments} that generalization improves when improving invariance to IDs.

\paragraph{Experimental setting}
We evaluate GraphConv \citep{morris2021weisfeiler}, GIN~\citep{gin}, and GAT~\citep{gat} with RNIs, on
four Open Graph Benchmark (OGB) datasets \citep{NEURIPS2020_fb60d411}, which cover node and graph classification tasks. 
For each dataset and model, we sampled $64$ random features as IDs. We compute the average \textit{invariance ratio} with respect to the training set $S$, defined as follows. 
 The invariance ratio of a model on an example $x$ is $\max_{i\in L} P[f(x)=i]$ where $P$ is the distribution over random ID assignments, and $L$ is the set of possible model outputs (e.g., classes). This measures the degree to which all random ID assignments result in the same label.
The maximum value of the invariance ratio is 1 (i.e., same output label for all random ID assignments) and the minimal value is $\nicefrac{1}{|L|}$.
For link prediction tasks, the invariance ratio is computed over the predicted binary edge labels (link or no-link), analogously to comparing predicted class labels in classification.
 We define the invariance ratio of $S$ as the average of the invariance ratios for all $x\in S$.
To estimate this ratio, we resampled IDs 10,000 times per sample after each epoch, on a random subset of up to 1,000 samples per dataset. All evaluations were performed in inference mode without gradient computation.

\paragraph{Datasets}
For node classification, we use ogbn-arxiv \citep{NEURIPS2020_fb60d411}, a citation network where nodes represent academic papers and edges denote citation relationships. Node features are derived from word embeddings of paper titles and abstracts, and the task involves predicting the subject area of each paper. For graph classification, we used three molecular property prediction datasets \citep{NEURIPS2020_fb60d411}. In ogbg-molhiv the task was to predict whether a molecule inhibits HIV replication, a binary classification task based on molecular graphs with atom-level features and bond-level edge features. ogbg-molbbbp involves predicting blood-brain barrier permeability, a crucial property for drug development, while ogbg-molbace focuses on predicting the ability of a molecule to bind to the BACE1 enzyme, associated with Alzheimer's disease. See more details in the Appendix.

\paragraph{Evaluation protocol}
We used the training-validation-test splits introduced in~\citet{NEURIPS2020_fb60d411}. Models were trained on the training set and hyperparameters were tuned by evaluating performance on the validation set. For testing, each model was trained with the best hyperparameters using three random seeds, and we report the average performance with these three seeds. Each network was trained for $500$ epochs. Detailed hyperparameters for each model and dataset are provided in the Appendix.

\paragraph{Results}
Figure~\ref{fig:invariance_curves} presents the invariance ratios for the different GNNs and datasets. The results indicate that the models did not achieve invariance with respect to the IDs. Notably, most invariance ratios continue to decrease rather than increase throughout the training process. This suggests that the GNNs not only fail to converge toward invariance but also tend to overfit as training progresses. 

We observe that the invariance ratios are relatively high at the beginning of training. It is important to note that a model can achieve invariance simply by ignoring the IDs (i.e., not using them at any part of the computation). As training progresses, the invariance ratio drops, which suggests that the model in fact uses the IDs, but in a non-invariant manner. This suggests that reassignment of random IDs during training is insufficient to guarantee invariance. In the next section, we theoretically analyze GNNs with IDs and consider approaches to achieve invariance.

We repeat this experiment with additional dimension sizes of random features and large weight decay and observe the same trends. Due to space limitations, these results are deferred to the Appendix.

\begin{table*}[t]
\captionsetup{type=figure}
    \centering

    \setlength{\tabcolsep}{3pt}
    \begin{tabular}{>{\centering\arraybackslash}m{0.03\textwidth}>{\centering}m{0.2175\textwidth}>{\centering\arraybackslash}m{0.2175\textwidth}>{\centering\arraybackslash}m{0.2175\textwidth}>{\centering\arraybackslash}m{0.2175\textwidth}}

        \toprule
        & {ogbn-arxiv} & {ogbg-molhiv} & {ogbg-molbbbp} & {ogbg-molbace} \\
        \midrule
        \rotatebox{90}{{GraphConv}} &
        \includegraphics[width=\linewidth]{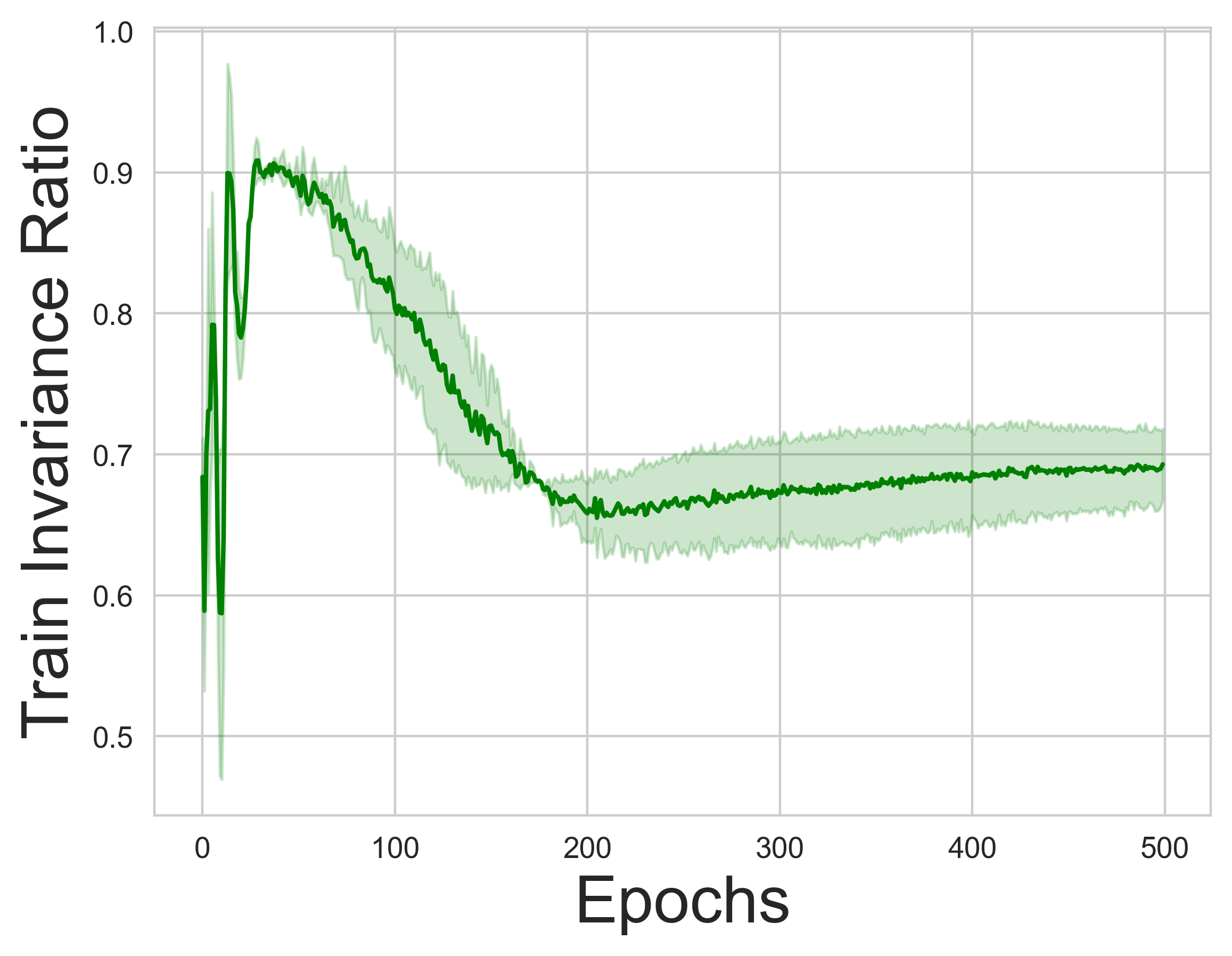} &
        \includegraphics[width=\linewidth]{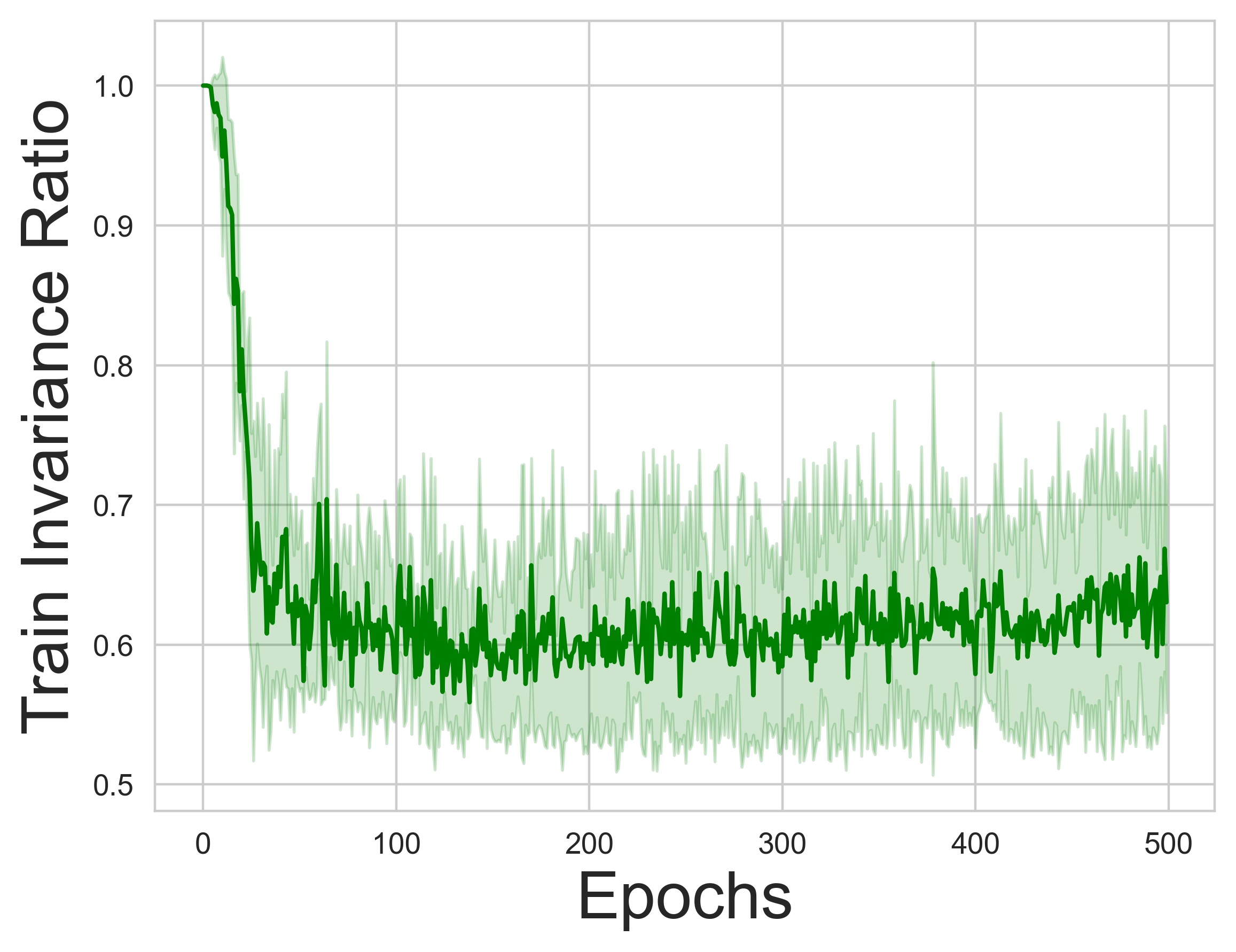} &
        \includegraphics[width=\linewidth]{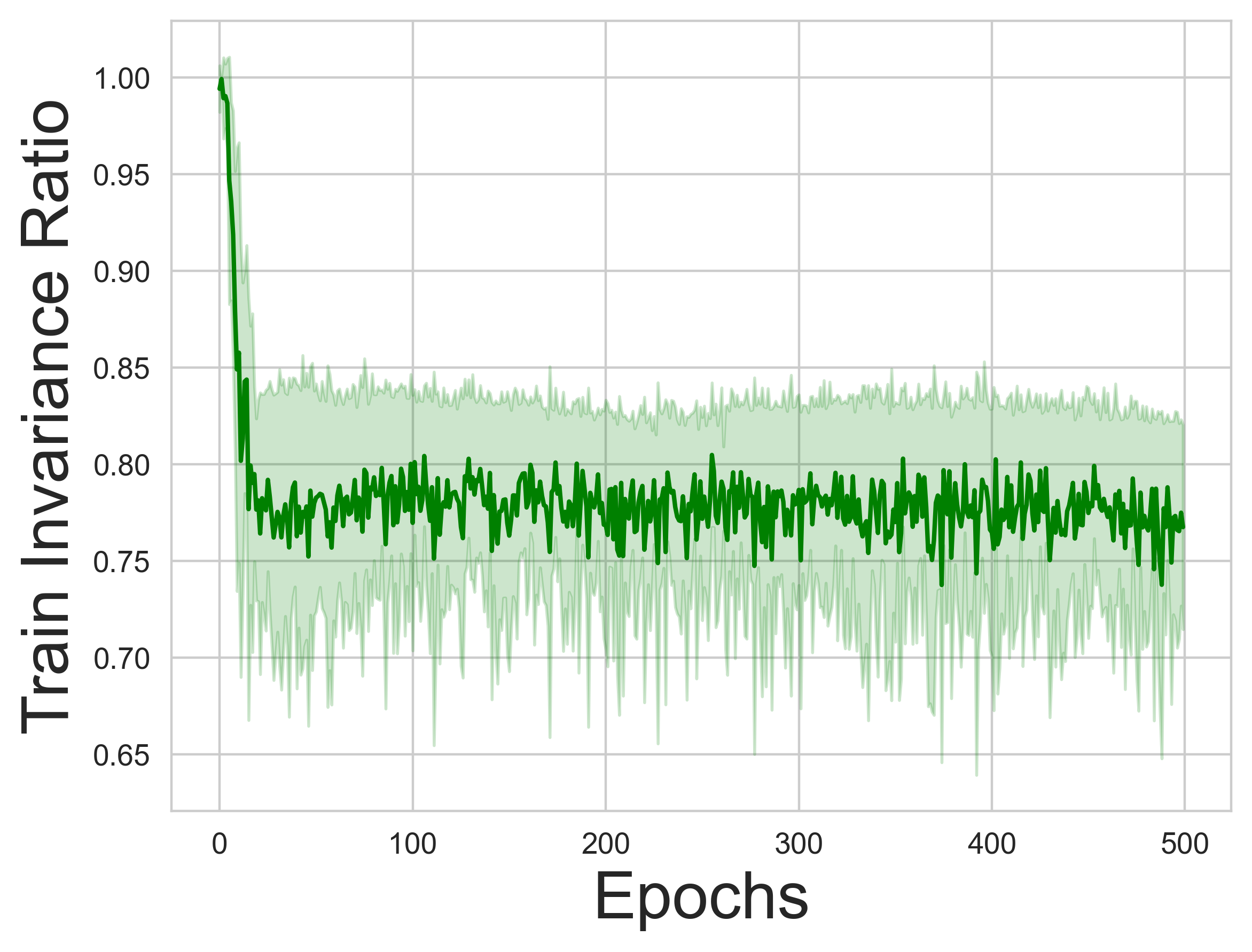} &
        \includegraphics[width=\linewidth]{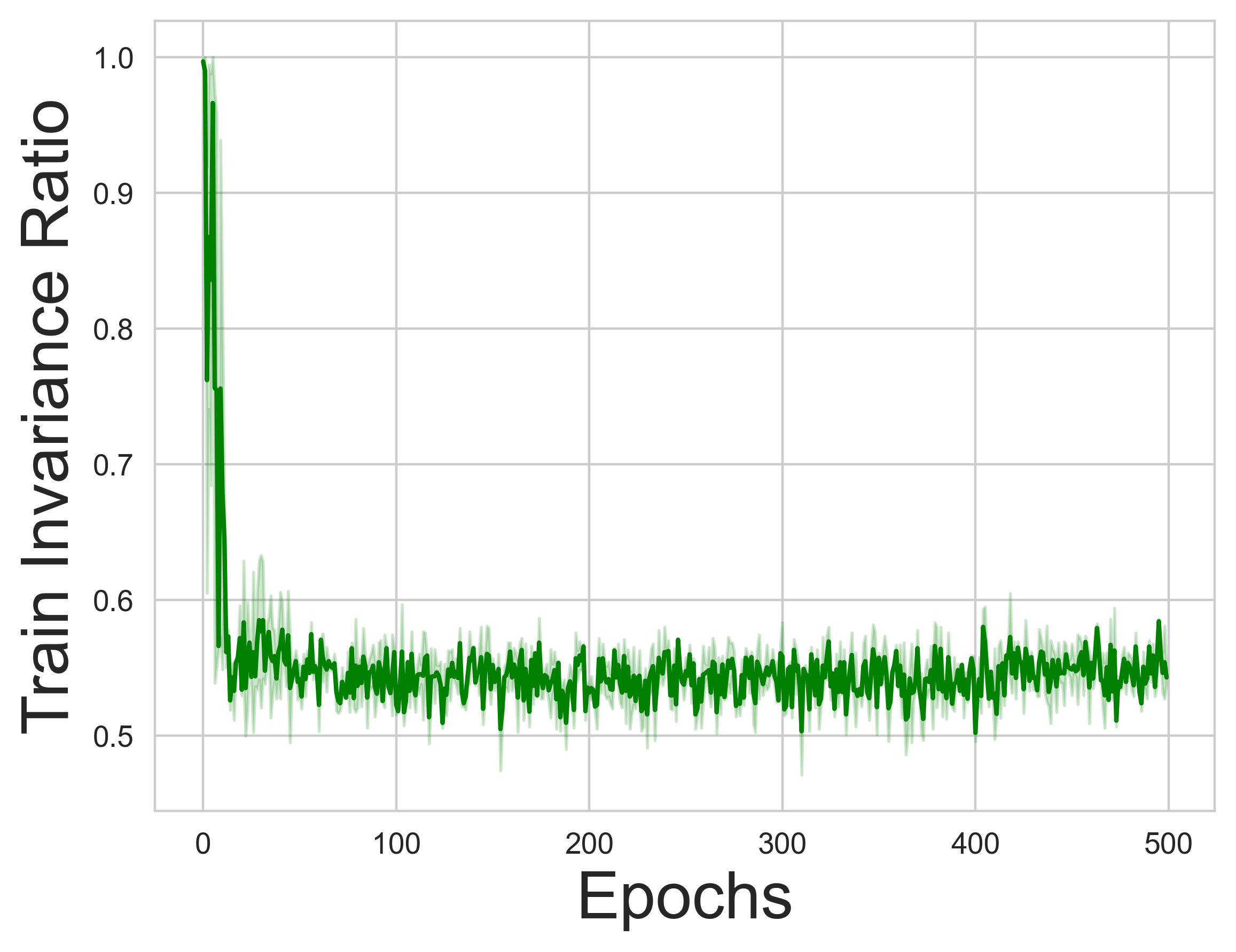} \\
        \midrule
        \rotatebox{90}{{GIN}} &
        \includegraphics[width=\linewidth]{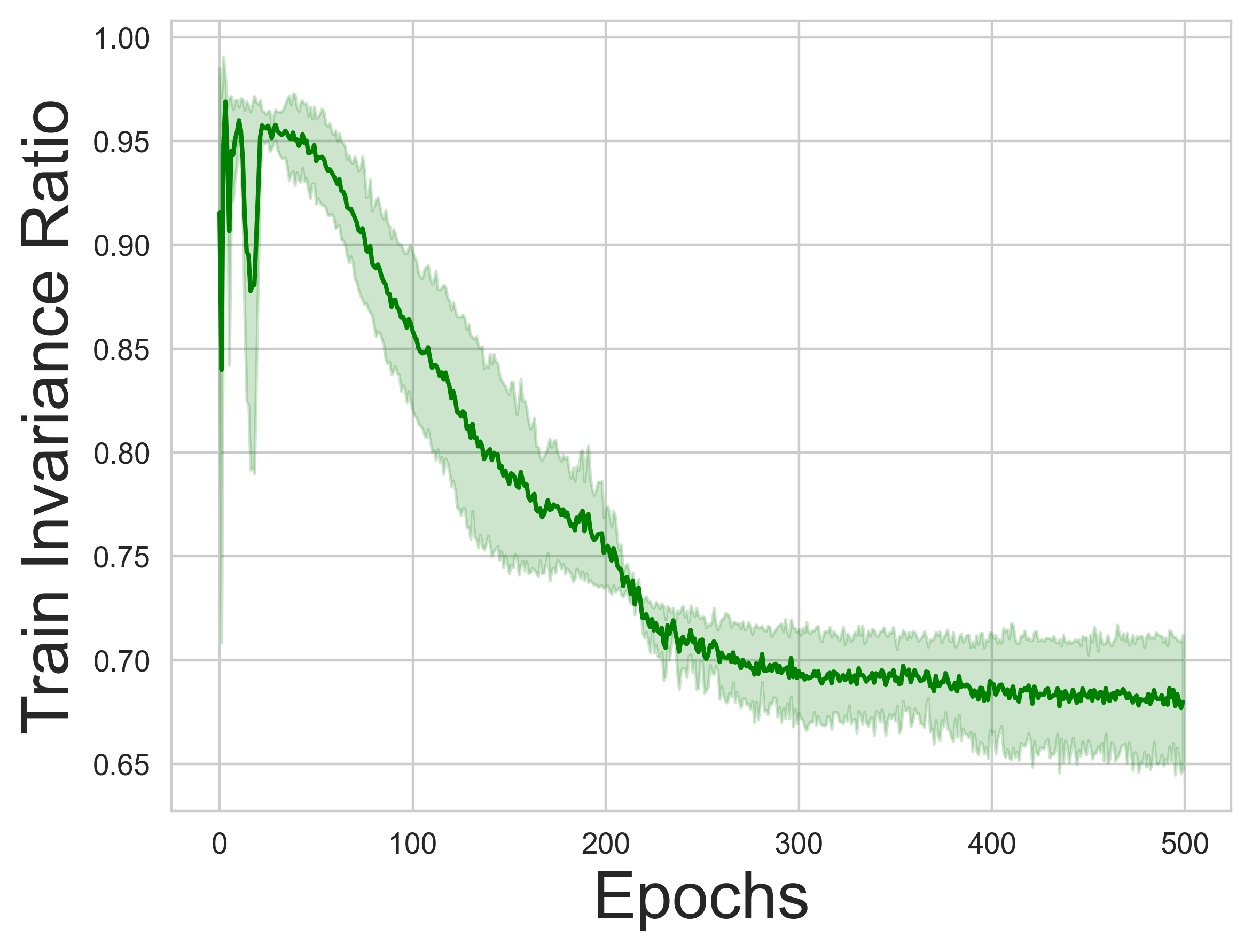} &
        \includegraphics[width=\linewidth]{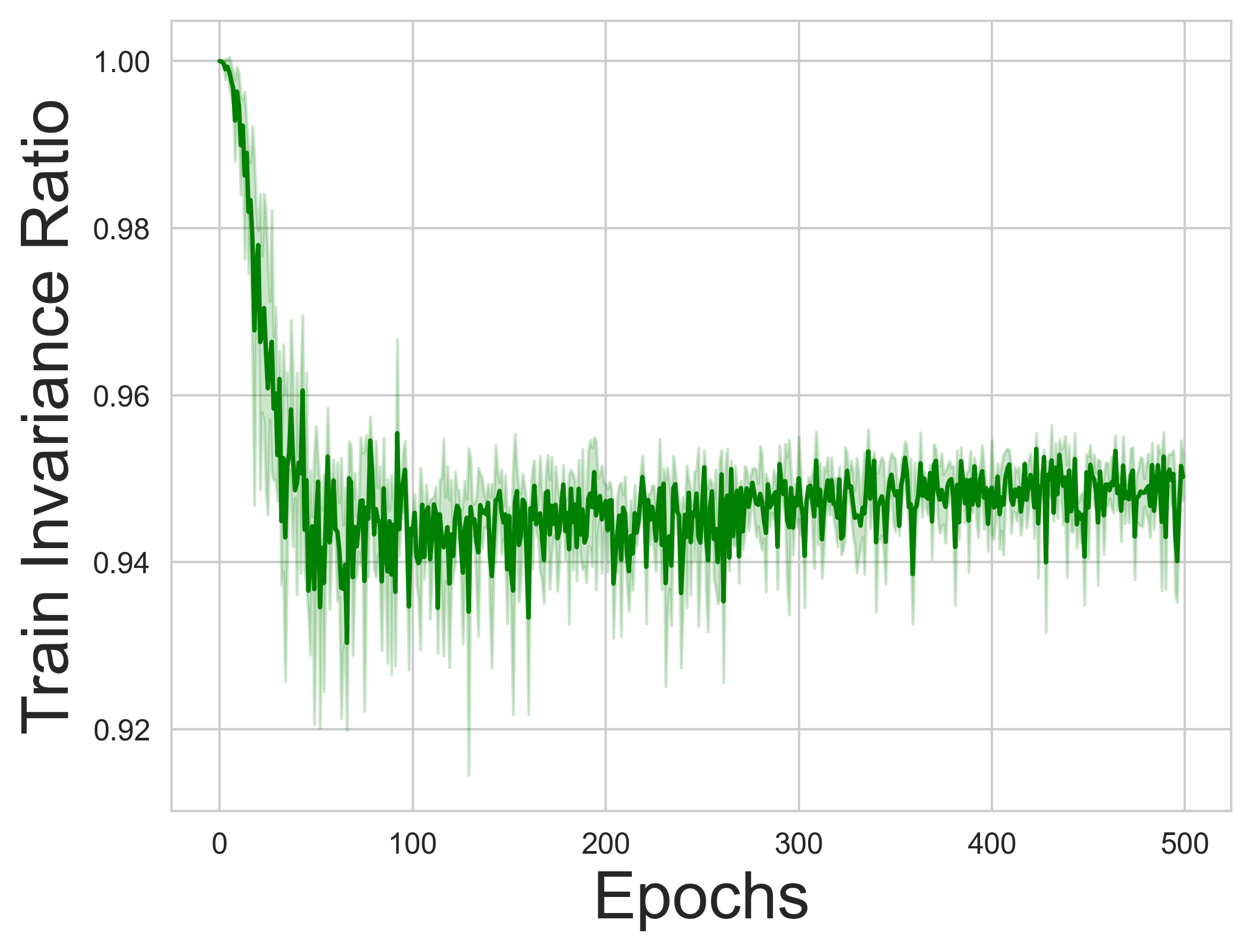} &
        \includegraphics[width=\linewidth]{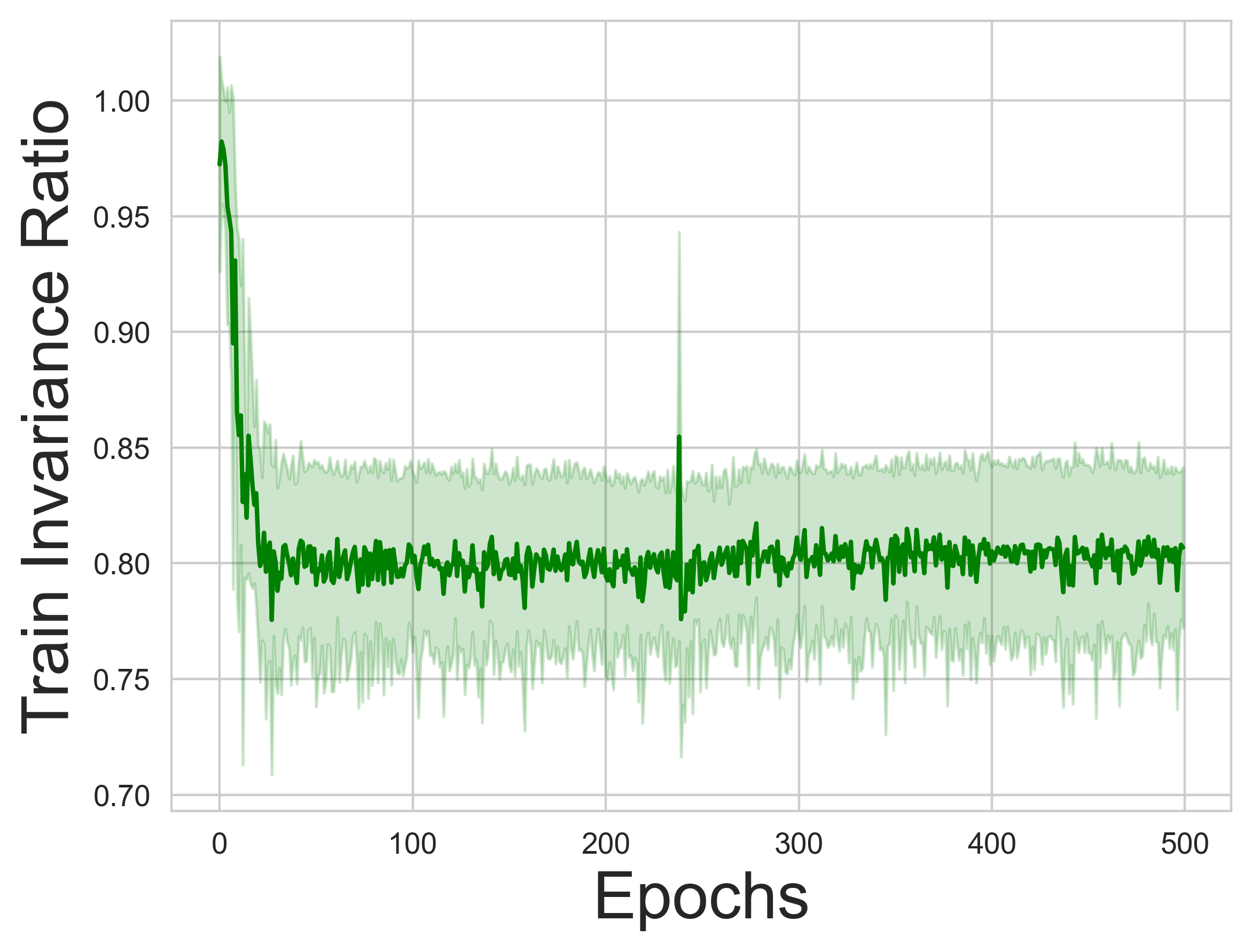} &
        \includegraphics[width=\linewidth]{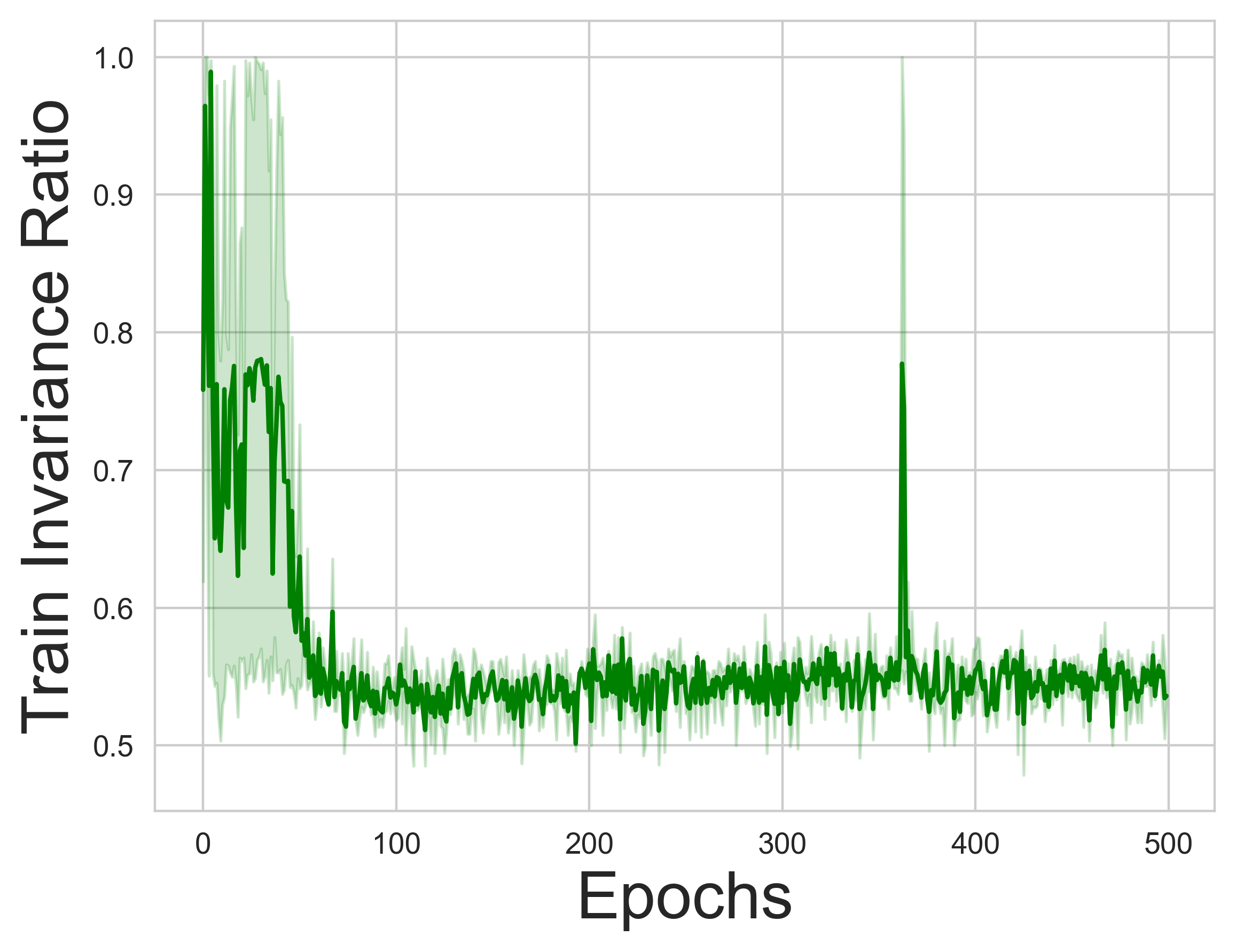} \\
        \midrule
        \rotatebox{90}{{GAT}} &
        \includegraphics[width=\linewidth]{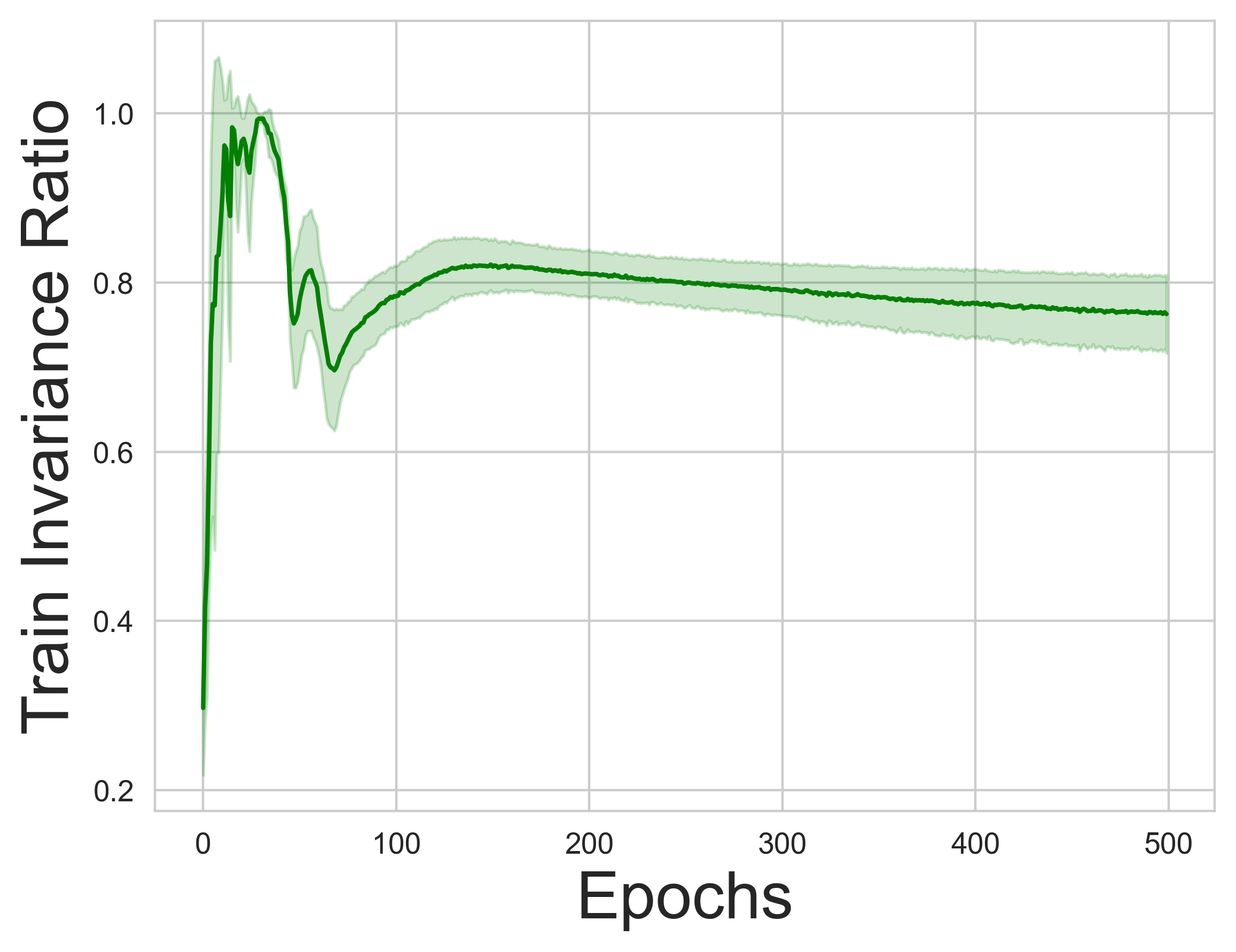} &
        \includegraphics[width=\linewidth]{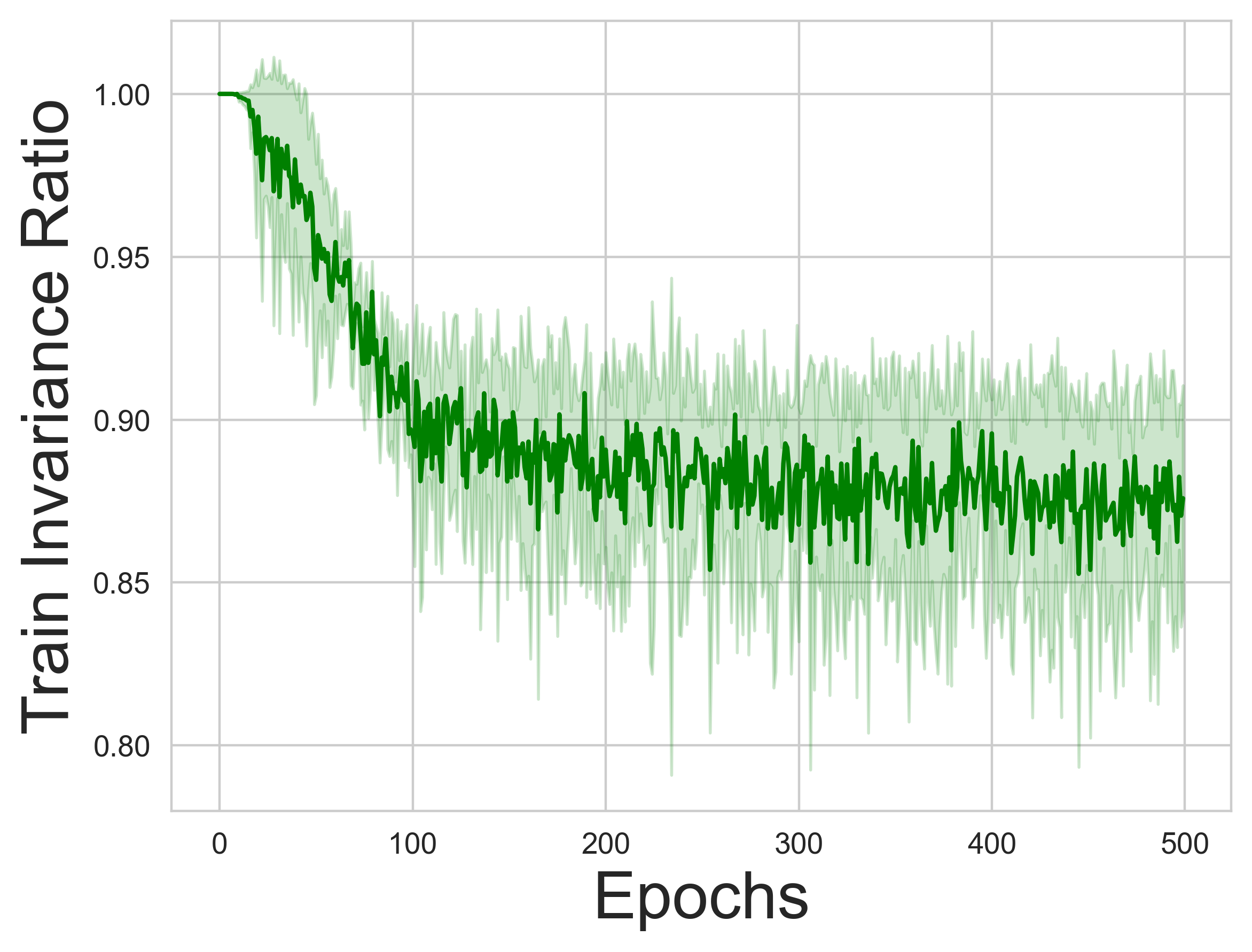} &
        \includegraphics[width=\linewidth]{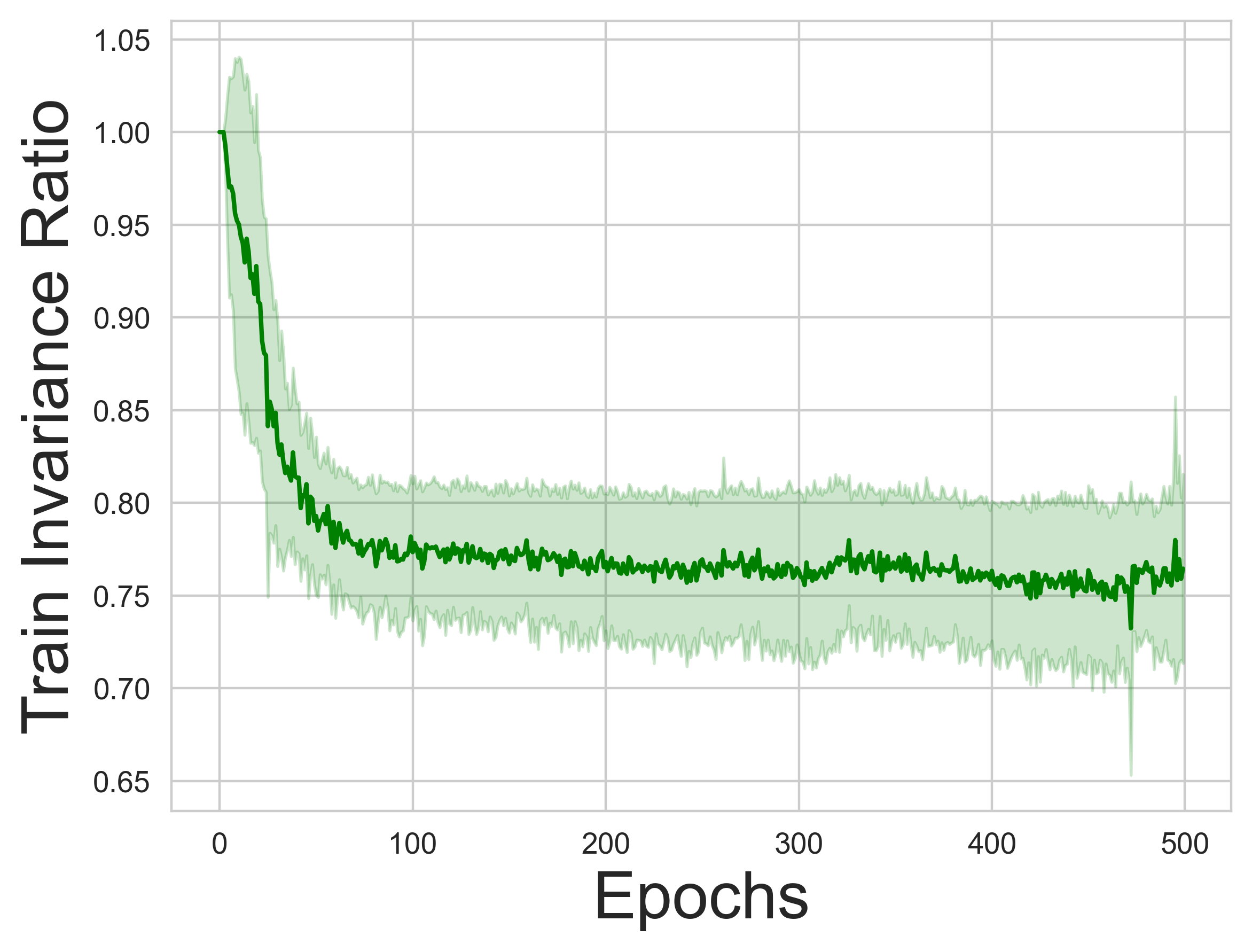} &
        \includegraphics[width=\linewidth]{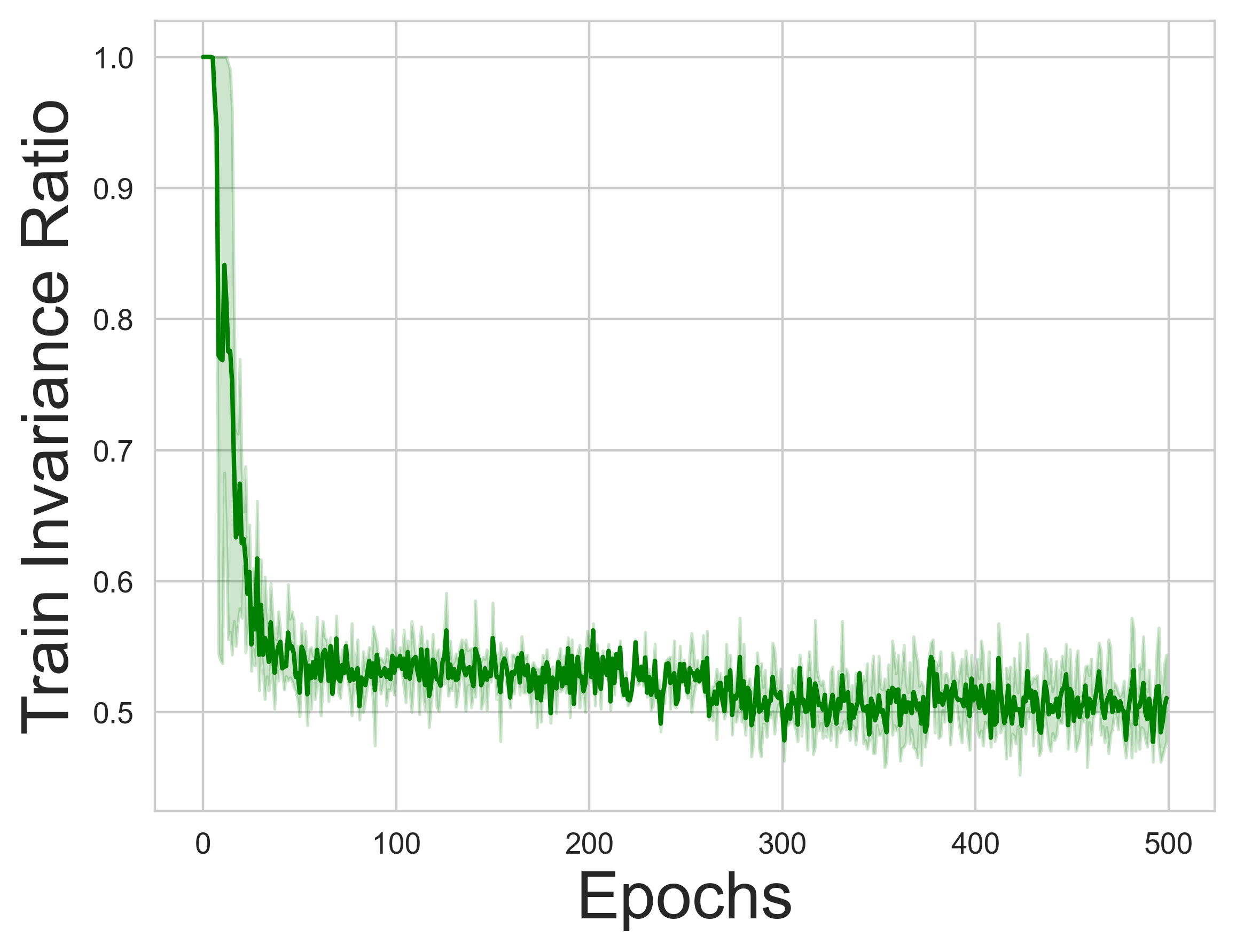} \\
        \bottomrule
    \end{tabular}
            \caption{Invariance ratios over different GNNs with RNI, on real-world datasets. Despite the ID resampling that is applied in every epoch, invariance is not achieved and, in most cases, does not increase during training.}
    \label{fig:invariance_curves}
\end{table*}

\section{Theoretical analysis of invariance to IDs and expressivity}
\label{sec:theory}

In this section, we establish theoretical results on GNNs with unique IDs. We examine how  GNNs can be invariant to IDs while still leveraging their expressiveness potential.
Our objective is to derive practical guidelines for introducing ID invariance into the model. Due to space constraints, all proofs are deferred to the Appendix.

\paragraph{Preliminaries} Throughout this paper, we denote a graph by $G = (V, E)$ over $n$ nodes and its corresponding node features by ${X} \in \mathbb{R}^{n\times d}$, where \( d \) is the input feature dimension.
We assume that a constant feature $1$ is always attached to the nodes, which is a common practice~\citep{gin, morris2021weisfeiler}.
We consider message-passing GNNs ~\cite{morris2021weisfeiler} defined as follows.
The initial node representations $H^{(0)}$ of a graph $G$ over $n$ nodes are the input node features $X$. Then,
each layer of the GNN updates node representations via:
\begin{equation*}
    \label{eq:general_gnn}
{H}^{(l+1)} = \textsc{GNN}_{\theta^{(l)}}({H}^{(l)}; G),
\end{equation*}
where $\textsc{GNN}_{\theta^{(l)}}$ is the $l$-th layer, and it depends on the input graph structure $G$ and the latest node representations ${H}^{(l-1)}$. 
The layer is associated with a set of learnable weights \( \theta^{(l)} \) at each layer \( l \in \{0, 1, \dots, L-1\} \).
The final node representations $\{h_i^{(L)}\}_{i=1}^{n}$ obtained in the last layer can then be used for downstream tasks such as node or graph labeling.
We denote the embedding function realized by the $k$'th layer of a GNN by $GNN_k$.

Our focus is on GNNs that use additional \textit{node IDs}. Namely, each node also takes an additional $r$ dimensional vector, which serves as its ID. Let the matrix  $I \in \mathbb{R}^{n \times r}$ denote the set of node IDs for all nodes.
Node IDs are simply concatenated to the input node features, so that $H^{(0)} \in \mathbb{R}^{n \times (d+r)}$.
We denote such GNNs as \textit{GNN-R}, and note that they take three inputs: a graph $G$, features $X$, and I, a matrix of node IDs. Their output is the output label. 

As noted earlier, a key desirable property of such models is that their output is invariant to the IDs, as formally defined below.
\begin{definition}[ID-invariant function]
Let $f$ be a GNN-R model, and $G$ be a graph. Then $f$ is \textit{ID-invariant} with respect to $G$ if for all $X$ and unique node IDs $I,I'$, $f(G,X,I) = f(G,X,I')$.
In words, the output of $f$ on $G$ with any node features $X$ does not depend on the values of the node IDs $I$. Similarly, $f$ is invariant with respect to a set of graphs $S$, if the definition above holds for all $G\in S$.
\end{definition}

We shall be specifically interested in the case where layers of a GNN are invariant to the IDs. Thus, we say that the $k$'th layer of a GNN is ID invariant if $GNN_k(G;X;I)$ is ID invariant. Namely, the layer is invariant if the embedding it produces does not change when $I$ changes. 

Note that a GNN-R that is not ID-invariant is prone to error, because there are certain inputs $X$ for which two different IDs $I,I'$ will result in different labels, and one of these will necessarily be wrong.
Therefore, ID-invariant solutions are preferred. Note also that a function may be ID-invariant with respect to some graphs, but non-ID-invariant with respect to others, as the following theorem shows:
\begin{theorem}\label{thm:invarinace_can_be_bad}
Let $S$ and $S'$ be sets of graphs such that $S' \setminus S$ contains a graph with at least one node. Then there exists a function $f$ that is ID-invariant with respect to $S$ and non-ID-invariant with respect to $S'$.
\end{theorem}
To prove Theorem~\ref{thm:invarinace_can_be_bad} we simply construct a function where ID-invariance is constrained on a set of graphs.
A direct implication of Theorem~\ref{thm:invarinace_can_be_bad} is that it is possible for a GNN-R to be ID-invariant on a training set, but non-invariant on a test set, leading to potentially high test error.

Since achieving invariance is desirable, we next ask how this can be accomplished. A natural idea is to enforce invariance at every layer. However, Theorem \ref{thm:mpnnwithidsepxressivity}  below shows that doing so actually eliminates the benefits of using IDs, reducing the model's expressiveness to that of a standard GNN without IDs. This is somewhat counterintuitive. It suggests that for expressiveness to be improved in the presence of unique IDs, at least some of the layers of the GNN must, in fact, be sensitive to the values of the IDs.

\begin{theorem}\label{thm:mpnnwithidsepxressivity}
Let $\mathbb{G}$ be a GNN without unique node IDs.
Let $\mathbb{G}'$ be a GNN-R with the same parameterization as $\mathbb{G}$, that is also ID-invariant in every layer. Then the sets of functions realized by $\mathbb{G}$ and $\mathbb{G}'$ are the same.
\end{theorem}

To prove Theorem~\ref{thm:mpnnwithidsepxressivity} we take a GNN-R and a GNN without IDs with the same parameterization and show by induction that the output of every layer of the GNN-R can be realized by the corresponding layers in the GNN without IDs (the reverse containment is clear).

Theorem~\ref{thm:mpnnwithidsepxressivity} implies that in order to design a network that is both ID-invariant and expressive, we must allow it to {\em not} be invariant to IDs in at least one hidden layer.
The next theorem shows that for three-layer models, invariance in the third layer results in enhanced expressiveness over a model that does not use IDs.
\begin{theorem}\label{thm:only_last_layer}
There exists a function $f(G;X)$ over graphs that no message passing GNN without IDs can realize, but $f(G;X)$ can be realized by an ID-invariant GNN-R with $3$ layers, where only the last layer is ID-invariant.
\end{theorem}

To prove Theorem~\ref{thm:only_last_layer} we construct an ID-invariant solution for the task of detecting a triangle. We show that it can be implemented by passing IDs in the first two layers, and then only checking if two IDs are the same or not, which is invariant to their specific values.

Together, Theorems~\ref{thm:mpnnwithidsepxressivity} and~\ref{thm:only_last_layer} answer not only \emph{whether} to enforce ID-invariance, but \emph{where} to enforce it: regularizing all layers toward invariance forfeits the expressive advantage of IDs, while regularizing only the final layer is sufficient for ID-invariance and compatible with expressiveness beyond 1-WL. This directly motivates the design of \ourmethod{} in Section~\ref{sec:method}.

In the next section, we build upon the results discussed in this section to design an approach that enhances the utilization of IDs in GNNs by explicitly enforcing ID-invariance to the model.

\section{Learning invariance to IDs}
\label{sec:method}

In this section, we build on our analysis in Section~\ref{sec:theory} and propose \ourmethod, an approach to improve the ability of GNNs to use IDs.
While regularizing toward invariance is a natural idea, the question remains \emph{how} to regularize: Theorem~\ref{thm:mpnnwithidsepxressivity} shows that requiring invariance in all layers eliminates the expressive benefit of IDs, whereas Theorem~\ref{thm:only_last_layer} shows that enforcing invariance solely at the final layer is sufficient for ID-invariance while preserving beyond-1-WL expressiveness.
Based on these insights, we propose to explicitly enforce invariance at the last layer only, using a regularizer. By avoiding invariance constraints on intermediate layers, \ourmethod~ maintains the model's capacity to leverage IDs effectively.

The key idea in {\ourmethod} is to add a term that regularizes models towards invariance to the random ID values. Before describing the regularizer term, we recall the ``standard'' supervised task-loss term.

Given a downstream task, such as graph classification or regression, we compute the task loss $ \mathcal{L}_{\text{task}}$ based on the output of the final GNN layer ${H}^{(L)}$. Specifically, we use a standard linear classifier $g: \mathbb{R}^{d} \rightarrow \mathbb{R}^{d_{\rm{out}}}$, where ${d_{\rm{out}}}$ is the desired dimension of the output (e.g., number of classes), to obtain the final prediction $\hat{y} = g({H}^{(L)})$. This output is fed into a standard supervised loss (e.g., mean-squared error or cross-entropy). Formally, the task loss reads:
\begin{equation*}
    \label{eq:task_loss}
\mathcal{L}_{\text{task}} = \text{Loss}( \hat{y}, y), 
\end{equation*}
where \( y \) is the ground truth label.

\begin{table*}[t]
\fontsize{7}{7.5}\selectfont
\centering
\setlength{\tabcolsep}{0.25em}
\caption{Invariance ratios for train and test sets when training GNNs with RNI vs. ICON. Across
all models and datasets, ICON consistently achieves high invariance ratios, sometimes reaching an
invariance ratio of 1. In contrast, RNI often yields significantly lower invariance ratios, in some cases
approaching 0.5 which is the lowest possible value for binary classification tasks.}
\label{tab:invariance_ogb}
\begin{tabular}{ll *{10}c}
\toprule
 &  & \multicolumn{2}{c}{GraphConv} & \multicolumn{2}{c}{GIN} & \multicolumn{2}{c}{GAT} & \multicolumn{2}{c}{TransformerConv} & \multicolumn{2}{c}{GraphGPS} \\
\cmidrule(lr){3-4} \cmidrule(lr){5-6} \cmidrule(lr){7-8} \cmidrule(lr){9-10} \cmidrule(lr){11-12}
DATASET & SPLIT & RNI & \ourmethod & RNI & \ourmethod & RNI & \ourmethod & RNI & \ourmethod & RNI & \ourmethod \\
\toprule
\multirow{2}{*}{\textsc{arxiv}} 
 & Train & 0.69$_{\pm0.02}$ & \textbf{0.79$_{\pm0.05}$} & 0.68$_{\pm0.03}$ & \textbf{1.00$_{\pm0.00}$} & 0.76$_{\pm0.05}$ & \textbf{0.90$_{\pm0.01}$} & 0.73$_{\pm0.04}$ & \textbf{0.94$_{\pm0.01}$} & 0.78$_{\pm0.03}$ & \textbf{0.98$_{\pm0.00}$} \\
 & Test  & 0.67$_{\pm0.05}$ & \textbf{0.85$_{\pm0.11}$} & 0.63$_{\pm0.02}$ & \textbf{1.00$_{\pm0.00}$} & 0.66$_{\pm0.03}$ & \textbf{0.85$_{\pm0.01}$} & 0.70$_{\pm0.03}$ & \textbf{0.91$_{\pm0.02}$} & 0.75$_{\pm0.03}$ & \textbf{0.97$_{\pm0.01}$} \\
\cmidrule(lr){1-12}
\multirow{2}{*}{\textsc{molhiv}} 
 & Train & 0.64$_{\pm0.10}$ & \textbf{0.95$_{\pm0.00}$} & 0.95$_{\pm0.00}$ & \textbf{1.00$_{\pm0.00}$} & 0.88$_{\pm0.04}$ & \textbf{0.94$_{\pm0.00}$} & 0.79$_{\pm0.02}$ & \textbf{0.97$_{\pm0.00}$} & 0.82$_{\pm0.02}$ & \textbf{0.99$_{\pm0.00}$} \\
 & Test  & 0.62$_{\pm0.09}$ & \textbf{0.98$_{\pm0.00}$} & 0.97$_{\pm0.00}$ & \textbf{1.00$_{\pm0.00}$} & 0.86$_{\pm0.03}$ & \textbf{0.96$_{\pm0.00}$} & 0.76$_{\pm0.03}$ & \textbf{0.98$_{\pm0.00}$} & 0.80$_{\pm0.02}$ & \textbf{1.00$_{\pm0.00}$} \\
\cmidrule(lr){1-12}
\multirow{2}{*}{\textsc{molbace}} 
 & Train & 0.76$_{\pm0.02}$ & \textbf{0.82$_{\pm0.01}$} & 0.78$_{\pm0.00}$ & \textbf{0.85$_{\pm0.11}$} & 0.74$_{\pm0.02}$ & \textbf{0.80$_{\pm0.01}$} & 0.70$_{\pm0.03}$ & \textbf{0.83$_{\pm0.02}$} & 0.74$_{\pm0.02}$ & \textbf{0.88$_{\pm0.02}$} \\
 & Test  & 0.74$_{\pm0.08}$ & \textbf{0.90$_{\pm0.03}$} & 0.82$_{\pm0.03}$ & \textbf{0.85$_{\pm0.14}$} & 0.72$_{\pm0.04}$ & \textbf{0.86$_{\pm0.02}$} & 0.68$_{\pm0.02}$ & \textbf{0.85$_{\pm0.01}$} & 0.73$_{\pm0.03}$ & \textbf{0.91$_{\pm0.02}$} \\
\cmidrule(lr){1-12}
\multirow{2}{*}{\textsc{molbbbp}} 
 & Train & 0.55$_{\pm0.00}$ & \textbf{0.52$_{\pm0.06}$} & 0.55$_{\pm0.02}$ & \textbf{0.93$_{\pm0.07}$} & 0.51$_{\pm0.03}$ & \textbf{0.58$_{\pm0.02}$} & 0.50$_{\pm0.02}$ & \textbf{0.66$_{\pm0.04}$} & 0.52$_{\pm0.01}$ & \textbf{0.73$_{\pm0.03}$} \\
 & Test  & 0.54$_{\pm0.01}$ & \textbf{0.67$_{\pm0.08}$} & 0.51$_{\pm0.04}$ & \textbf{0.96$_{\pm0.04}$} & 0.56$_{\pm0.02}$ & \textbf{0.60$_{\pm0.07}$} & 0.53$_{\pm0.01}$ & \textbf{0.70$_{\pm0.03}$} & 0.55$_{\pm0.02}$ & \textbf{0.76$_{\pm0.01}$} \\
\cmidrule(lr){1-12}
\multirow{2}{*}{\textsc{collab}} 
 & Train & 0.61$_{\pm0.03}$ & \textbf{0.89$_{\pm0.02}$} & 0.62$_{\pm0.03}$ & \textbf{0.97$_{\pm0.01}$} & 0.63$_{\pm0.03}$ & \textbf{0.90$_{\pm0.02}$} & 0.65$_{\pm0.03}$ & \textbf{0.93$_{\pm0.02}$} & 0.66$_{\pm0.02}$ & \textbf{0.97$_{\pm0.01}$} \\
 & Test  & 0.59$_{\pm0.04}$ & \textbf{0.92$_{\pm0.03}$} & 0.60$_{\pm0.02}$ & \textbf{0.98$_{\pm0.01}$} & 0.61$_{\pm0.02}$ & \textbf{0.92$_{\pm0.02}$} & 0.62$_{\pm0.02}$ & \textbf{0.95$_{\pm0.02}$} & 0.64$_{\pm0.03}$ & \textbf{0.98$_{\pm0.01}$} \\
\cmidrule(lr){1-12}
\multirow{2}{*}{\textsc{squirrel}} 
 & Train & 0.57$_{\pm0.02}$ & \textbf{0.83$_{\pm0.04}$} & 0.58$_{\pm0.03}$ & \textbf{0.91$_{\pm0.03}$} & 0.59$_{\pm0.02}$ & \textbf{0.84$_{\pm0.03}$} & 0.60$_{\pm0.02}$ & \textbf{0.87$_{\pm0.02}$} & 0.61$_{\pm0.02}$ & \textbf{0.92$_{\pm0.02}$} \\
 & Test  & 0.56$_{\pm0.03}$ & \textbf{0.87$_{\pm0.02}$} & 0.55$_{\pm0.04}$ & \textbf{0.94$_{\pm0.02}$} & 0.58$_{\pm0.03}$ & \textbf{0.86$_{\pm0.03}$} & 0.59$_{\pm0.02}$ & \textbf{0.89$_{\pm0.02}$} & 0.60$_{\pm0.02}$ & \textbf{0.95$_{\pm0.02}$} \\
\bottomrule
\end{tabular}
\end{table*}

\begin{table*}[h!]
\fontsize{7.5}{8.5}\selectfont
\renewcommand{\arraystretch}{0.95}
\centering
\setlength{\tabcolsep}{0.35em}
\caption{Performance across OGB datasets for different models. We report mean ROC-AUC and std for the molhiv, molbace and molbbbp
datasets,  mean accuracy and std for the arxiv and Squirrel dataset, and Hits@50 and std for collab.
For each model we report the performance with no IDs, with RNI, and with \ourmethod. The best-performing approach within the std margin is highlighted in bold. In all cases, \ourmethod~outperforms RNI. }
\label{tab:final_accuracy}
\begin{tabular}{lccccccc}
\toprule
\textbf{MODEL} & \textsc{arxiv} & \textsc{molhiv} & \textsc{molbace} & \textsc{molbbbp} & \textsc{collab} & \textsc{squirrel} \\
\midrule
GraphConv                & 72.05$_{\pm 1.13}$ & \textbf{70.27$_{\pm 2.51}$}  & 62.58$_{\pm 1.98}$ & 63.39$_{\pm 0.17}$ & 46.15$_{\pm 1.01}$ & 32.4$_{\pm 1.7}$ \\
GraphConv+RNI            & 71.34$_{\pm 1.50}$ & 56.44$_{\pm 2.29}$ & 52.29$_{\pm 2.01}$ & 59.08$_{\pm 2.03}$ & 40.09$_{\pm 3.49}$ & 28.9$_{\pm 2.2}$ \\
GraphConv+\ourmethod     & \textbf{73.66$_{\pm 1.19}$} & 68.59$_{\pm 1.96}$ & \textbf{65.72$_{\pm 1.10}$} & \textbf{67.02$_{\pm 0.97}$} & \textbf{48.15$_{\pm 1.71}$} & \textbf{34.8$_{\pm 1.3}$} \\
\midrule
GIN                      & 76.30$_{\pm 1.58}$ & \textbf{72.09$_{\pm 1.01}$} & 73.02$_{\pm 1.27}$ & \textbf{60.86$_{\pm 1.05}$} & 43.87$_{\pm 0.94}$ & 30.2$_{\pm 1.5}$ \\
GIN+RNI                  & 71.25$_{\pm 1.91}$ & 71.06$_{\pm 1.95}$ & 61.94$_{\pm 2.51}$ & 53.35$_{\pm 1.73}$ & 39.42$_{\pm 2.31}$ & 26.7$_{\pm 1.9}$ \\
GIN+\ourmethod           & \textbf{79.16$_{\pm 0.82}$} & \textbf{72.01$_{\pm 1.30}$} & \textbf{76.78$_{\pm 0.99}$} & \textbf{59.96$_{\pm 2.00}$} & \textbf{45.76$_{\pm 1.12}$} & \textbf{33.1$_{\pm 1.1}$} \\
\midrule
GAT                      & \textbf{72.76$_{\pm 1.30}$} & 72.60$_{\pm 2.49}$ & \textbf{71.01$_{\pm 2.83}$} & \textbf{59.42$_{\pm 2.52}$} & 44.23$_{\pm 1.23}$ & 29.8$_{\pm 2.0}$ \\
GAT+RNI                  & 61.85$_{\pm 3.11}$ & 54.52$_{\pm 2.10}$ & 63.43$_{\pm 1.39}$ & 58.75$_{\pm 1.98}$ & 40.87$_{\pm 1.98}$ & 27.3$_{\pm 1.6}$ \\
GAT+\ourmethod           & \textbf{71.77$_{\pm 2.96}$} & \textbf{76.60$_{\pm 1.18}$} & 66.08$_{\pm 1.35}$ & \textbf{61.93$_{\pm 1.29}$} & \textbf{45.02$_{\pm 1.35}$} & \textbf{32.7$_{\pm 1.2}$} \\
\midrule
TransformerConv          & 73.15$_{\pm 1.39}$ & \textbf{70.11$_{\pm 1.49}$} & 74.91$_{\pm 2.04}$ & \textbf{60.92$_{\pm 1.37}$} & 44.67$_{\pm 1.10}$ & 33.6$_{\pm 1.4}$ \\
TransformerConv+RNI      & 70.62$_{\pm 1.60}$ & 67.92$_{\pm 1.40}$ & 65.33$_{\pm 1.98}$ & 55.38$_{\pm 1.73}$ & 42.21$_{\pm 1.92}$ & 30.9$_{\pm 1.6}$ \\
TransformerConv+\ourmethod & \textbf{76.09$_{\pm 2.10}$} & \textbf{73.25$_{\pm 1.10}$} & \textbf{75.01$_{\pm 1.50}$} & \textbf{59.18$_{\pm 1.82}$} & \textbf{46.39$_{\pm 1.18}$} & \textbf{35.3$_{\pm 1.3}$} \\
\midrule
GraphGPS                 & 74.80$_{\pm 1.99}$ & \textbf{72.75$_{\pm 2.11}$} & 69.94$_{\pm 1.51}$ & \textbf{58.17$_{\pm 2.01}$} & 47.12$_{\pm 0.88}$ & \textbf{37.4$_{\pm 1.2}$} \\
GraphGPS+RNI             & 71.03$_{\pm 2.08}$ & 61.15$_{\pm 3.79}$ & 68.18$_{\pm 1.40}$ & 56.93$_{\pm 2.40}$ & 44.90$_{\pm 1.30}$ & 33.7$_{\pm 1.7}$ \\
GraphGPS+\ourmethod      & \textbf{77.25$_{\pm 2.15}$} & \textbf{74.19$_{\pm 1.72}$} & \textbf{72.01$_{\pm 1.49}$} & \textbf{60.35$_{\pm 1.72}$} & \textbf{48.26$_{\pm 0.91}$} & 36.6$_{\pm 1.0}$ \\
\bottomrule
\end{tabular}
\end{table*}

To achieve invariance to the IDs, {\ourmethod} utilizes a regularizer 
$\mathcal{L}_{\text{reg}}$ that is computed as follows.
Assume our training sample is \( G_1 = (V, E, {X}) \) with node identifiers $R_1$. 
We sample additional random node IDs $R_2$. Our goal is to make the network
output similar for these two random IDs. Namely, the model should be invariant to the specific values of the node IDs. Let \( {H}_i^{(L)} \) denote the final embeddings for the model with node IDs $R_i$. Namely  \( {H}_i^{(L)} \) is obtained by propagating the inputs with IDs $R_i$ through the network up to the last GNN layer (prior to the final prediction layer).
 To impose ID-invariance, we consider a loss that aims to reduce the difference between these embeddings. There are several options and metrics possible here, and we choose the mean squared error for simplicity. Namely, we consider the regularizer:
\begin{equation*}
\label{eq:residual_loss}\mathcal{L}_{\text{reg}} =  \frac{1}{n \cdot d_L}\|{H}_1^{(L)}- {H}_2^{(L)}\|_2^2,
\end{equation*}
where $n$ is the number of nodes and $d_L$ is the embedding dimension, so that $\mathcal{L}_{\text{reg}}$ is normalized and does not scale with the graph size.

Then, the overall loss per training example in \ourmethod~ is:
\(
    \mathcal{L} = \mathcal{L}_{\text{task}} +  \mathcal{L}_{\text{reg}}.
\) \newline
The task loss $\mathcal{L}_{\text{task}}$ is computed using the predictions from the first ID assignment $R_1$ only, while $\mathcal{L}_{\text{reg}}$ compares the embeddings from both assignments.
The loss is minimized via standard stochastic gradient descent, where this is the per-sample loss. We note that no additional weighting coefficient is needed between the two terms, as the normalization in $\mathcal{L}_{\text{reg}}$ ensures that both terms are on a comparable scale.
Note that it is possible to sample more than two random IDs per training example and adapt the loss accordingly. Here we opt for the simplest version where the sample regularizer only uses two samples.

\section{Experimental evaluation}
\label{sec:experiments}

In this section, we evaluate \ourmethod{} on diverse real-world  and synthetic datasets. We demonstrate its ability to consistently improve invariance to IDs and, as a result, to often enhance generalization.

\subsection{Real-world datasets}\label{subsec:real_data}
To assess the effectiveness of {\ourmethod} on real datasets, we evaluate it on diverse tasks from the Open Graph Benchmark (OGB)~\citep{NEURIPS2020_fb60d411}. We test {\ourmethod} across multiple GNN architectures, including GraphConv \citep{morris2021weisfeiler}, GIN \citep{gin}, GAT \citep{gat}, TransformerConv \citep{ijcai2021p214} and GraphGPS \citep{NEURIPS2022_5d4834a1}.  We note that {\ourmethod} is simple to add to any existing method. We focus on these architectures due to their widespread use and to demonstrate the efficacy of {\ourmethod} across GNN architectures.

\textbf{Datasets} As node-level tasks we used ogbn-arxiv and the heterophilous Squirrel dataset for node classification. We used ogbg-molhiv, ogbg-molbbbp, and ogbg-molbace for binary graph classification, and ogbl-collab for link prediction \citep{NEURIPS2020_fb60d411, 10.1093/comnet/cnab014}.
The datasets are described in detail with their dataset statistics in the Appendix.

\begin{wraptable}{r}{0.4\textwidth}
\vspace{-10pt}
\fontsize{8}{9}\selectfont
\setlength{\tabcolsep}{4pt}
\centering
\caption{Invariance ratios with respect to the train set and test set when training GNNs with RNI vs. with \ourmethod{} over the isInTriangle task. \ourmethod~ achieves perfect invariance.}
\label{tab:is_triangle_invariance}
\begin{tabular}{lcc}
\\
\toprule 
  \textbf{METHOD}   & Train & Test \\
  \toprule
GraphConv+RNI    &    0.95$_{\pm 0.01}$ & 0.93$_{\pm 0.08}$\\ 
GraphConv+\ourmethod   & \textbf{1.00$_{\pm 0.00}$} & \textbf{1.00$_{\pm 0.00}$}\\ 
\midrule
GIN+RNI    &    0.95$_{\pm 0.02}$ & 0.94$_{\pm 0.02}$\\
GIN+\ourmethod  & \textbf{1.00$_{\pm 0.00}$} & \textbf{0.99$_{\pm 0.00}$}\\
\midrule
GAT+RNI    &   0.88$_{\pm 0.09}$ & 0.88$_{\pm 0.13}$\\
GAT+\ourmethod  &\textbf{1.00$_{\pm 0.00}$} &  \textbf{1.00$_{\pm 0.00}$}\\ 
\bottomrule
\end{tabular}
\vspace{-5pt}
\end{wraptable}

\textbf{Setup}
For each dataset and GNN, we evaluated three configurations: the baseline model without IDs, the model with RNI, and the model with \ourmethod. 
For each configuration,  we conducted a grid search by training on the training set and evaluating on the validation set. We selected the best-performing model over the validation set. We then report the average metric scores and std of the selected configuration with $3$ seeds. We use ROC-AUC for binary graph classification, Accuracy for node classification, and Hits@50 for link prediction, following the common metrics for these datasets in the literature.
We also calculate the final invariance ratios as defined in Section~\ref{sec:overfitting} for both RNI and \ourmethod.
We use $64$ random features as the IDs.
We repeat this experiment with additional sizes of ID dimensions and observe the same trends. Due to space limitations, these results are deferred to the Appendix.

\begin{wraptable}{r}{0.4\textwidth}
\centering
   \vspace{-0.8em}
\fontsize{7}{8}\selectfont
\caption{Accuracy on the isInTriangle task in interpolation and extrapolation settings.}
\label{tab:int_vs_exp}
\setlength{\tabcolsep}{3pt}
\renewcommand{\arraystretch}{1.05}
\begin{tabular}{lcc}
\toprule
\textbf{Method} & Interp. & Extrap. \\
\midrule
GraphConv+Const.       & 75.35$_{\pm 2.09}$ & 53.70$_{\pm 1.67}$ \\
GraphConv+RNI          & 74.87$_{\pm 3.06}$ & 57.02$_{\pm 3.39}$ \\ 
GraphConv+\ourmethod{} & \textbf{88.45$_{\pm 2.04}$} & \textbf{78.20$_{\pm 2.53}$} \\ 
\midrule
GIN+Const.             & 76.10$_{\pm 1.83}$ & 56.29$_{\pm 1.55}$ \\
GIN+RNI                & 72.15$_{\pm 2.01}$ & 55.01$_{\pm 1.30}$ \\
GIN+\ourmethod{}       & \textbf{90.95$_{\pm 1.27}$} & \textbf{71.20$_{\pm 1.94}$} \\
\midrule
GAT+Const.             & 74.99$_{\pm 2.10}$ & 56.07$_{\pm 2.11}$ \\
GAT+RNI                & 71.20$_{\pm 1.48}$ & 58.92$_{\pm 1.57}$ \\
GAT+\ourmethod{}       & \textbf{89.71$_{\pm 1.27}$} & 79.30$_{\pm 1.90}$ \\
\bottomrule

\end{tabular}
    \vspace{-1em}
\end{wraptable}
\textbf{Results} Table~\ref{tab:invariance_ogb} presents the invariance ratios of \ourmethod~ and RNI across all datasets and models, with respect to the train and test sets. Across all datasets and models, \ourmethod~ significantly improves invariance with respect to both the train and the test set. In some cases, it reaches full invariance. 
Table~\ref{tab:final_accuracy} presents the average ROC-AUC scores for each model and dataset. Across all models and datasets, \ourmethod~ improves upon RNI, with the maximal margin being 22.08 ROC-AUC points on the ogbg-molhiv datasets with GAT. In $23$ out of the $30$ cases, \ourmethod~improved generalization also upon the baseline model and was on par with the baseline in $4$ other cases, which shows that improving invariance can improve generalization. 

\subsection{Synthetic datasets}

We next examine the performance of \ourmethod~ and RNI on synthetic datasets with three synthetic tasks where GNNs without IDs are provably unable to solve these tasks \citep{NEURIPS2020_75877cb7, pmlr-v119-garg20c, ijcai2021p291} but GNNs that employ unique IDs and are sufficiently large can solve these tasks \citep{ijcai2021p291}.

\subsubsection{The ``Is In Triangle'' task}

\textbf{Dataset}
The isInTriangle task is a binary node classification where the goal is to determine whether a given node is part of a triangle. It was shown in~\citet{NEURIPS2020_75877cb7, pmlr-v119-garg20c} that MPGNNs without IDs cannot solve the isInTriangle task. With IDs, GNNs can solve the task, yet they can do so by either overfitting to the IDs or with an IDs-invariant solution. We show the existence of an IDs-invariant solution for this task in the proof of Theorem~\ref{thm:only_last_layer}.
The dataset consists of 100 graphs with 100 nodes each, generated using the preferential attachment (BA) model \citep{badist}, in which graphs are constructed by incrementally adding new nodes with $m$ edges and connecting them to existing nodes with a probability proportional to the degrees of those nodes. We adopt an inductive setting, where the graphs used during testing are not included in the training set. 
We used $m=2$ for the training graphs and evaluated two test sets: an interpolation setting in which the graphs are drawn from the BA distribution with $m=2$ and an extrapolation setting in which the graphs are drawn from a different distribution than the train graphs, a BA distribution with $m=3$.

\textbf{Setup} We repeat the setup from Sections~\ref{sec:overfitting} and \ref{subsec:real_data}.
As in this task nodes have no features, for the baseline models where IDs are not used, we use a constant 1 feature for all nodes.
For each task, we conducted a grid search by training on the training set and evaluating on the validation set. We selected the best-performing model over the validation set.
We then report the average accuracy and std of the selected configuration with $3$ seeds.
We also calculate the final invariance ratios as in Sections~\ref{sec:overfitting} and \ref{subsec:real_data} for RNI and {\ourmethod}.

\textbf{Results}
Table~\ref{tab:is_triangle_invariance} presents the invariance ratios of \ourmethod~ and RNI across all models, for the interpolation and extrapolation settings. Across all models, \ourmethod{} substantially improves invariance both with respect to the train and the test set. In some cases, it reaches full invariance. 
Table~\ref{tab:int_vs_exp} presents the average accuracy for each model for the interpolation and extrapolation setup. Across all models \ourmethod~ improves upon RNI and the baseline, with the maximal margin being 18.80 accuracy percentage in the interpolation setting and 21.18 accuracy percentage in the extrapolation setting.

\subsubsection{EXP and CEXP}

\citet{ijcai2021p291}
presented the EXP and CEXP datasets as an example of the improved expressiveness RNI provides, as these tasks cannot be solved with MPGNNs without IDs.
 \begin{wrapfigure}{r}{0.36\textwidth}
    \centering
    \vspace{-0.5em}
    \includegraphics[width=0.36\textwidth]{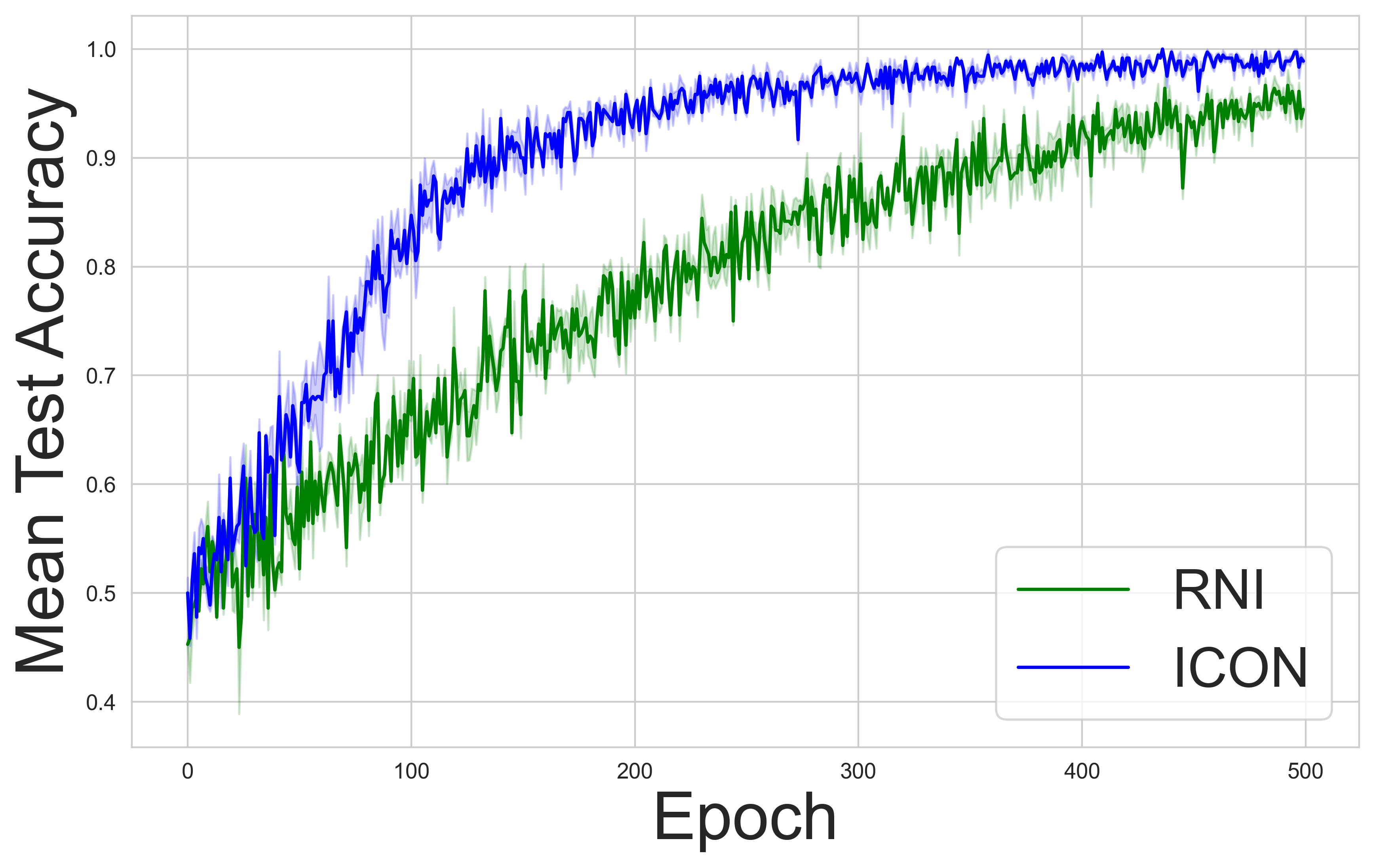}
    \includegraphics[width=0.36\textwidth]{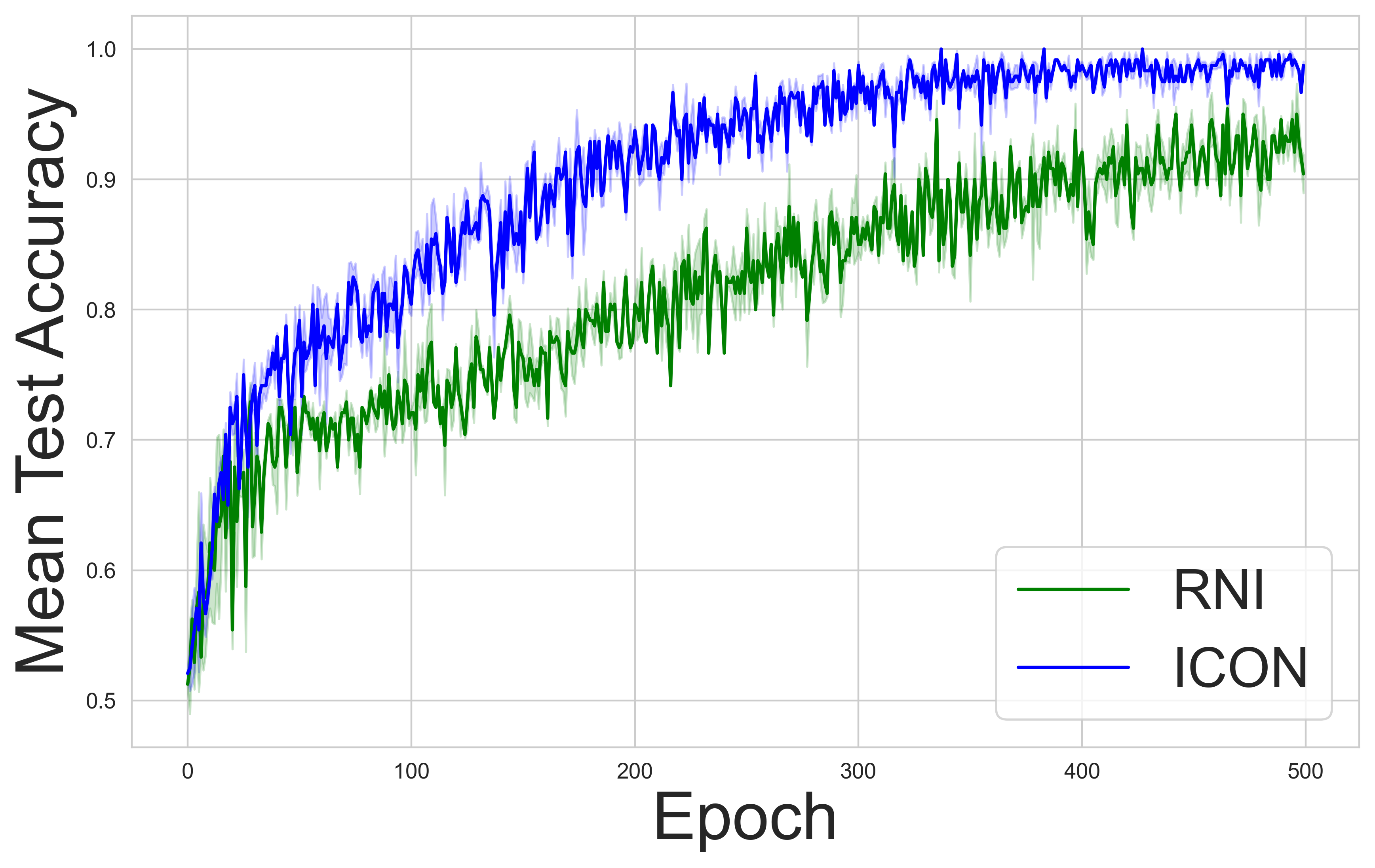}

    \caption{The test accuracy learning curves of RNI and \ourmethod~on the EXP and CEXP datasets.}
    \label{Figure:exp_cexp}
\end{wrapfigure}
However, it was shown that to improve upon the MPGNNs accuracy, much longer training time is required. In the current experiment, we set out to test if {\ourmethod} can achieve the same accuracy of RNI while improving the training time.
The datasets contain 600 pairs of graphs that cannot be distinguished by 1\&2-WL tests. The goal is to classify each graph into one of two isomorphism classes. Splitting is done by stratified five-fold cross-validation. CEXP is a modified version of EXP, where 50\% of pairs are slightly modified to be distinguishable by 1-WL. It was shown in \citet{ijcai2021p291} that a GraphConv \citep{ijcai2021p291} with RNI reaches perfect accuracy on the test set.
We repeat the protocol from \citet{ijcai2021p291}, who evaluated RNI. We use a GraphConv GNN~\citep{morris2021weisfeiler} with RNI and \ourmethod~with $8$ layers, $64$ random features, $64$ hidden dimensions, over 500 epochs, with $3$ random seeds.
Figure~\ref{Figure:exp_cexp} presents the learning curves of accuracy during training for \ourmethod~and RNI. Both methods reach a test accuracy of almost 100\%, but \ourmethod~convergence is faster in both tasks, highlighting the effect of explicit regularization towards IDs-invariance as introduced by \ourmethod~ vs. an implicit one as done in RNI using random resampling of IDs during training.

\section{Conclusions}
In this paper, we harness the theoretical power of unique node identifiers (IDs) to enhance the effectiveness of Graph Neural Networks (GNNs). We show that, in practice, the final output of GNNs that incorporate node IDs is not invariant to the specific choice of IDs.
We provide a theoretical analysis that motivates an approach that explicitly enforces invariance to IDs in GNNs while taking advantage of their expressive power. Building on our theoretical results, we propose \ourmethod{}, an explicit regularization technique, which is GNN agnostic.
 Through comprehensive experiments, we demonstrate across real-world and synthetic datasets that \ourmethod{} considerably improves invariance to node IDs, as well as generalization, extrapolation, and training convergence time. These results highlight \ourmethod{} as an efficient solution to improve the performance of GNNs through the utilization of IDs, making it a valuable tool for use with any GNN.
We discuss directions for future work in Appendix~\ref{app:future}.

\acksection
This work was supported by the Israeli Science Foundation grants 2247/23.

\bibliographystyle{plainnat}
\bibliography{biblio}

\appendix

\section{Proofs}
\subsection{Proof of Theorem~\ref{thm:invarinace_can_be_bad}}
Let \(S\) and \(S'\) be sets of graphs such that \(S' \setminus S\) contains a graph with at least one node.
Fix any ID-invariant graph property \(P : \mathcal{G} \to \{0,1\}\) (e.g., ``\(G\) has an Eulerian path''). 
For node features \(X\) and an assignment \(I\) that gives each node \(v \in V(G)\) a unique ID vector \(I_v \in \mathbb{R}^{r}\), define \(f(G,X,I)\), ignoring \(X\), as
\[
f(G,X,I) \;:=\;
\begin{cases}
P(G), & \text{if } G \in S,\\[4pt]
\displaystyle \sum_{v \in V(G)} \mathbf{1}^\top I_v, & \text{if } G \in S' \setminus S,\\[4pt]
0, & \text{otherwise},
\end{cases}
\]
where \(\mathbf{1} \in \mathbb{R}^{r}\) is the all-ones vector.

\emph{ID-invariance on \(S\):} If \(G \in S\), then \(f(G,X,I) = P(G)\) depends only on \(G\) and is independent of \(X\) and \(I\). Hence \(f\) is ID-invariant on \(S\).

\emph{Non-invariance on \(S'\):} Suppose \(G \in S' \setminus S\) with \(V(G) \neq \varnothing\) (which exists by hypothesis).
Pick any node \(u \in V(G)\) and any two ID assignments \(I, I'\) that are identical on \(V(G) \setminus \{u\}\) but satisfy \(\mathbf{1}^\top I_u \neq \mathbf{1}^\top I'_u\) (e.g., \(I'_u = I_u + \varepsilon e_1\) for \(\varepsilon \neq 0\) small enough to avoid the finitely many values taken by other nodes, which preserves uniqueness and changes the sum). Then
\[
f(G,X,I) - f(G,X,I')
= \sum_{v \in V(G)} \mathbf{1}^\top I_v - \sum_{v \in V(G)} \mathbf{1}^\top I'_v
= \mathbf{1}^\top I_u - \mathbf{1}^\top I'_u \;\neq\; 0,
\]
for any fixed \(X\), so \(f(G,X,I) \neq f(G,X,I')\). Hence \(f\) is not ID-invariant on \(S'\).

Therefore a single function \(f\) is ID-invariant on \(S\) and not ID-invariant on \(S'\).

\subsection{Proof of direct implication of Theorem~\ref{thm:invarinace_can_be_bad}}
In the main text we mention that a direct implication of Theorem~\ref{thm:invarinace_can_be_bad} states that for finite disjoint train and test sets, a GNN-R can be ID-invariant on the train set and non-ID-invariant on the test set.
This setting only differs from the setting in Theorem~\ref{thm:invarinace_can_be_bad} in that the learned function is now restricted to one that can be realized by a GNN.
Indeed, as proven in~\citet{ijcai2021p291}, a GNN with IDs is a universal approximator on finite collections of graphs. Specifically, with enough layers and width, an MPGNN can reconstruct the adjacency matrix in some hidden layer, after which membership in the finite train set reduces to finite lookup, while the ID-dependent branch is global sum pooling. It can therefore realize the same case distinction as in the proof of Theorem~\ref{thm:invarinace_can_be_bad}.

\subsection{Proof of Theorem~\ref{thm:mpnnwithidsepxressivity}}

Let $\mathbb{G}$ be a message-passing GNN as in the preliminaries, without node IDs.
Let $\mathbb{G}'$ be a GNN-R with node IDs $I \in \mathbb{R}^{n\times r}$ concatenated to the features, sharing the same architecture and parameterization as $\mathbb{G}$.
Let $g$ be a fixed graph.

We will show that if $\mathbb{G}'$ is ID-invariant in every layer, then $\mathbb{G}$ and $\mathbb{G}'$ can realize the same set of functions. Specifically, $\mathbb{G}'$ is not more expressive than $\mathbb{G}$.
We prove this by induction on the number of layers.

We first consider GNNs that output embeddings for each node in the graph. The argument for GNNs performing graph-level prediction (e.g., via aggregation and readout over node embeddings) follows immediately and is discussed at the end of the proof.
We assume, without loss of generality, that the hidden dimension $h$ is fixed across all layers, including the final node embeddings.

\underline{Base step: one-layer GNN.}
Let $\mathbb{G}'$ be a one-layer GNN-R and $\mathbb{G}$ a one-layer GNN with the same architecture and parameterization as $\mathbb{G}'$.
The inputs to the first layer are $(A, [X\;I])$ in $\mathbb{G}'$ and $(A, X)$ in $\mathbb{G}$.
Recall that both models include a constant feature $1$ as part of $X$ (see Section~\ref{sec:theory}).

\emph{$\mathbb{G}$ can realize anything $\mathbb{G}'$ can.}
Assume $\mathbb{G}'$ is ID-invariant and implements an ID-invariant function $f$.
In a message-passing layer, the update for each node $v$ takes the form
\[
h_v^{(1)} = u\!\Big(h_v^{(0)},\; \sum_{w \in N(v)} m(h_w^{(0)}, h_v^{(0)})\Big),
\]
where $m$ is the message function and $u$ is the update function.
Since each message $m(h_w^{(0)}, h_v^{(0)})$ is a function of a single neighbor-receiver pair, and the aggregation is a sum, it suffices to show that the pairwise message function does not depend on the ID components.

Since $f$ is ID-invariant on the entire graph, it is also ID-invariant when restricted to any induced subgraph. In particular, consider any edge $(w, v)$ and the induced subgraph on just these two nodes. On this subgraph, the message from $w$ to $v$ is:
\[
m([X_w; I_w], [X_v; I_v]) = m([X_w; I'_w], [X_v; I'_v])
\]
for all valid unique IDs $I, I'$. Since this holds for arbitrary ID values, the message function effectively depends only on $(X_w, X_v)$, i.e., there exists a function $\tilde{m}$ such that $m([X_w; I_w], [X_v; I_v]) = \tilde{m}(X_w, X_v)$ for all unique IDs. The same argument applies to the update function $u$ with respect to the self-features.

Since each individual message ignores the IDs, the full aggregation $\sum_{w \in N(v)} \tilde{m}(X_w, X_v)$ is a permutation-invariant function over the neighbor multiset that depends only on node features $X$, regardless of the neighborhood size. Therefore, $\mathbb{G}$ can realize the same function by setting its weights for the feature components equal to those of $\mathbb{G}'$.

\emph{$\mathbb{G}'$ can realize anything $\mathbb{G}$ can.}
Given any function realized by $\mathbb{G}$ operating on features $X$ (which include the constant $1$), $\mathbb{G}'$ receives the concatenated input $[X; I]$. Since both models share the same architecture and parameterization, $\mathbb{G}'$ can simply set the weights corresponding to the $I$ channels to zero, reducing to the same computation as $\mathbb{G}$.

\underline{Inductive step.}
Assume the theorem holds for any GNN and GNN-R with up to $L-1$ layers.
We now prove it for an $L$-layer GNN and GNN-R.

Let $\mathbb{G}'$ be an $L$-layer GNN-R and $\mathbb{G}$ an $L$-layer GNN with the same architecture and parameterization as $\mathbb{G}'$.
Let $f$ be a function implemented by $\mathbb{G}'$.
We can decompose it as $f = f_{L} \circ f_{[L-1]}$, where $f_{[L-1]}$ is the function implemented by the first $L-1$ layers of $\mathbb{G}'$, and $f_{L}$ is the function implemented by the $L$-th layer.

By the inductive hypothesis, $f_{[L-1]}$ can be realized by some $\mathbb{G}_{(L-1)}$ (a GNN without IDs) with the same architecture and parameterization up to layer $L-1$.
We therefore set $\mathbb{G}_{(L-1)}$ to implement $f_{[L-1]}$.
Then, since the architectures of $\mathbb{G}$ and $\mathbb{G}'$ are identical, $\mathbb{G}$ can realize the same $L$-th layer transformation $f_L$.
Hence, we define $\mathbb{G}$ such that its first $L-1$ layers realize $f_{[L-1]}$ and its $L$-th layer realizes $f_L$.

Consequently,
\[
\mathbb{G} = f_L \circ f_{[L-1]} = f,
\]
and thus $\mathbb{G}$ can realize $f$.

The reverse direction is the same:
let $\mathbb{G}$ be an $L$-layer GNN without IDs, and $\mathbb{G}'$ an $L$-layer GNN-R
with the same architecture and parameterization.
Assume that $\mathbb{G}$ realizes some function $f = f_{L} \circ f_{[L-1]}$,
where $f_{[L-1]}$ corresponds to its first $L-1$ layers and $f_{L}$ to its last layer.

By the inductive hypothesis, there exists a GNN-R $\mathbb{G}'_{(L-1)}$ (ID-invariant)
whose first $L-1$ layers realize $f_{[L-1]}$.
Since the architectures and parameters of $\mathbb{G}$ and $\mathbb{G}'$ are identical,
we can define the final layer of $\mathbb{G}'$ to implement the same transformation $f_L$.
Thus, the full $\mathbb{G}'$ realizes
\[
\mathbb{G}' = f_L \circ f_{[L-1]} = f.
\]
Therefore, any function realizable by $\mathbb{G}$ can also be realized by $\mathbb{G}'$,
which completes the inductive step and establishes the equivalence for $L$ layers.
\qed

\subsection{Proof of Theorem~\ref{thm:only_last_layer}}

It is known that \textsc{isInTriangle} is not realizable by MPGNNs without IDs ~\citep{pmlr-v119-garg20c} (i.e., not within 1-WL expressivity). 
We construct a 3-layer GNN-R $f$ that realizes \textsc{isInTriangle}, where only the last layer is ID-invariant. Crucially, the construction works for \emph{any} set of unique node IDs, not just a specific assignment.

Let $G=(V,E)$ with $|V|=n$. Each node $u\in V$ is assigned a unique scalar ID $I_u\in\mathbb{R}$ (no other features are used; the extension to $I_u \in \mathbb{R}^r$ is immediate by projecting to a scalar first).
Write $N(u)$ for the neighbors of $u$ and let $h_v^{(\ell)}$ denote the representation of node $v$ at layer $\ell$.

For the sake of clarity, we define the message-passing steps as commonly done in the literature~\citep{pmlr-v70-gilmer17a, Battaglia2018, gin, morris2021weisfeiler}.
A layer $\ell$ is specified by a message function $m^{(\ell)}$, an aggregation operator $\operatorname{AGG}$ (we use summation), and an update function $u^{(\ell)}$:
\[
\text{messages:}\quad M_{u\to v}^{(\ell)} \;=\; m^{(\ell)}\!\big(h_u^{(\ell-1)},\, h_v^{(\ell-1)}\big),\qquad
\text{aggregate:}\quad a_v^{(\ell)} \;=\; \sum_{u\in N(v)} M_{u\to v}^{(\ell)},
\]
\[
\text{update:}\quad h_v^{(\ell)} \;=\; u^{(\ell)}\!\big(h_v^{(\ell-1)},\, a_v^{(\ell)}\big).
\]

We now construct a three-layer GNN-R. The construction uses a collection of $K$ nonlinear basis functions $\psi_1, \ldots, \psi_K : \mathbb{R} \to \mathbb{R}$ (e.g., random Fourier features or shifts of a ReLU), chosen so that for any set of $n$ distinct scalars $\{I_1, \ldots, I_n\}$, the vectors $\Psi(I_u) \coloneqq (\psi_1(I_u), \ldots, \psi_K(I_u))^\top \in \mathbb{R}^K$ are linearly independent. Such a collection exists for $K \geq n$ by standard arguments (e.g., using analytic functions, the vectors are linearly independent for any set of distinct inputs). In particular, for distinct IDs, the inner product $\langle \Psi(I_u), \Psi(I_w) \rangle$ is large when $I_u = I_w$ and generically small otherwise, allowing the detection of shared neighbors.

We set $h_v^{(0)} \coloneqq \Psi(I_v) \in \mathbb{R}^K$.

\medskip
\noindent\textbf{Layer 1: neighbor signature vectors.}
Each node collects the nonlinear embeddings of its neighbors' IDs:
\[
m^{(1)}(h_u^{(0)}, h_v^{(0)}) \;=\; h_u^{(0)} = \Psi(I_u),\qquad
u^{(1)}(h_v^{(0)}, a_v^{(1)}) \;=\; a_v^{(1)}.
\]
Thus
\[
h_v^{(1)} \;=\; \sum_{u \in N(v)} \Psi(I_u) \;\in\; \mathbb{R}^K.
\]
This is a ``neighborhood signature'' vector that encodes the multiset of neighbor IDs. Note that this layer is \emph{not} ID-invariant: the representation $h_v^{(1)}$ depends on the specific ID values of $v$'s neighbors.

\medskip
\noindent\textbf{Layer 2: common-neighbor detection.}
Define the message as the inner product of the two endpoints' Layer-1 signatures:
\[
m^{(2)}(h_u^{(1)}, h_v^{(1)}) \;=\; \langle h_u^{(1)},\, h_v^{(1)} \rangle \;=\; \sum_{w \in N(u)} \sum_{w' \in N(v)} \langle \Psi(I_w),\, \Psi(I_{w'}) \rangle \;\in\; \mathbb{R},
\]
and the update
\[
u^{(2)}(h_v^{(1)}, a_v^{(2)}) \;=\; a_v^{(2)} \;=\; \sum_{u \in N(v)} \langle h_u^{(1)},\, h_v^{(1)} \rangle.
\]
By the linear independence of $\{\Psi(I_w)\}$, the Gram matrix $G_{ww'} = \langle \Psi(I_w), \Psi(I_{w'}) \rangle$ is positive definite. In particular, the diagonal entries satisfy $G_{ww} = \|\Psi(I_w)\|^2 > 0$, while the off-diagonal terms $G_{ww'}$ for $w \neq w'$ are generically nonzero but can be made arbitrarily small relative to the diagonal by choosing the basis functions $\psi_k$ appropriately (e.g., with sufficiently separated random frequencies).

Therefore, the dominant contribution to $h_v^{(2)}$ comes from terms where $w = w'$, i.e., common neighbors:
\[
h_v^{(2)} \;=\; \sum_{u \in N(v)} \sum_{w \in N(u)} \sum_{w' \in N(v)} G_{ww'} \;\approx\; c \cdot \sum_{u \in N(v)} |N(u) \cap N(v)|,
\]
where $c = \|\Psi(I_w)\|^2 > 0$ (which can be made constant across nodes by normalizing $\Psi$). With a suitable choice of basis functions, the gap between the dominant (common-neighbor) terms and the cross terms can be made arbitrarily large, so that $h_v^{(2)} > 0$ if and only if $v$ participates in at least one triangle.

We emphasize that the \emph{value} of $h_v^{(2)}$ depends on the specific IDs (through the Gram matrix), so this layer is not ID-invariant. However, the \emph{sign} of $h_v^{(2)}$ --- which is all that matters for the final decision --- is determined solely by the graph topology.

\medskip
\noindent\textbf{Layer 3: ID-invariant decision.}
Let $g:\mathbb{R}\to\{0,1\}$ be a function that realizes the threshold $g(t)=\mathbb{1}\{t>\tau\}$ for a threshold $\tau$ chosen between the cross-term noise level and the common-neighbor signal level (which is always possible by the gap argument above). This can be implemented by a ReLU MLP.
Define a dummy message,
\[
m^{(3)}(h_u^{(2)},h_v^{(2)}) \;=\; 0,\qquad
a_v^{(3)} \;=\; 0,
\]
and the update
\[
u^{(3)}(h_v^{(2)},a_v^{(3)}) \;=\; g\!\big(h_v^{(2)}\big)\in\{0,1\}.
\]
Set the network output to $f(G)_v \coloneqq h_v^{(3)}$.

\medskip
By the construction above,
\[
f(G)_v \;=\; 1 \quad\Longleftrightarrow\quad h_v^{(2)} \;>\; \tau
\quad\Longleftrightarrow\quad \exists\,u\in N(v)\ \text{with}\ N(u)\cap N(v)\neq\varnothing,
\]
which holds if and only if node $v$ lies on a triangle. Thus $f$ realizes \textsc{isInTriangle}.

\medskip
\noindent\textbf{ID-invariance of the construction.}
The output $f(G)_v \in \{0,1\}$ depends only on whether $v$ participates in a triangle, which is a topological property independent of the ID values. Hence, for any two sets of unique IDs $I$ and $I'$, the network produces the same output: $f(G, I) = f(G, I')$. Layers 1 and 2 produce ID-dependent intermediate representations (their numerical values change when IDs are resampled), but the final binary output is invariant. Only Layer 3 is constrained to be ID-invariant; Layers 1 and 2 are not.

\medskip
\noindent\textbf{Remark on the hidden dimension.}
The construction uses a hidden dimension $K \geq n$ (the number of nodes). While this depends on the graph size, this is standard in expressivity proofs for GNNs (cf.~\citep{ijcai2021p291}), which establish the existence of networks with sufficient capacity, not practical architectures.

\section{Additional Experimental Results}
\label{app:additional-results}

In this section, we provide complementary experiments to those presented in the main paper, aimed at verifying the robustness of the observed phenomena with respect to the number of random ID channels and to the regularization strength during training. The results consistently support the conclusions drawn in the main text regarding both the invariance and generalization benefits achieved by \textsc{ICON}.

\paragraph{Varying the number of random ID channels.}
\Cref{tab:wd} reports results obtained when using $8$ random channels as node identifiers (IDs), compared to the default configuration of $64$ channels used in the main experiments. We observe that the trends remain consistent across all models and datasets. Specifically, \textsc{ICON} continues to achieve substantially higher ID-invariance ratios than RNI and, in most cases, leads to improved generalization performance. While using fewer ID channels slightly reduces the overall representational capacity of the model, the relative improvement achieved by \textsc{ICON} remains stable, demonstrating its robustness to the dimensionality of the identifier space.

\begin{figure*}[t]
    \centering

    \includegraphics[width=1\textwidth]{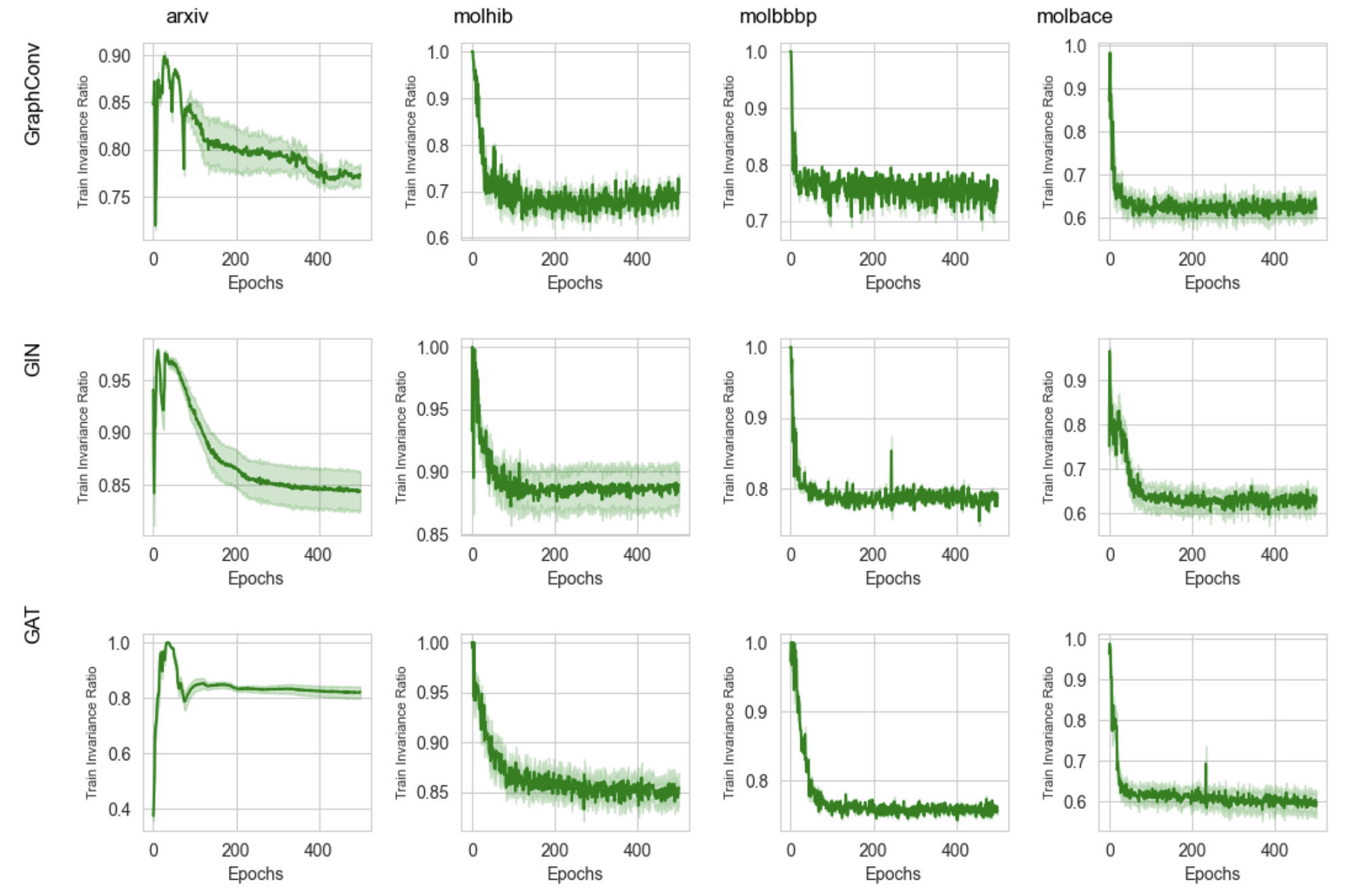} \\

    \caption{Invariance ratios over different GNNs with RNI, on real-world datasets, with weight decay 0.00001. Despite the ID resampling that is applied in every epoch, invariance is not achieved and, in most cases, does not increase during training.}
    \label{Figure:wd_inv}
\end{figure*}

\paragraph{Invariance under reduced weight decay.}
To further assess the stability of \textsc{ICON} under different regularization regimes, we repeat all experiments with a reduced weight decay coefficient of $10^{-5}$ (compared to $10^{-4}$ in the main experiments). The corresponding invariance ratios are illustrated in \Cref{Figure:wd_inv}. Once again, the results align with the findings in the main paper: models trained with \textsc{ICON} exhibit markedly higher invariance ratios throughout training, even when the weight decay is reduced. This suggests that the observed invariance improvement is a direct result of the \textsc{ICON} regularization term rather than a by-product of stronger implicit regularization.

\begin{figure*}[t]
    \centering

    \includegraphics[width=1\textwidth]{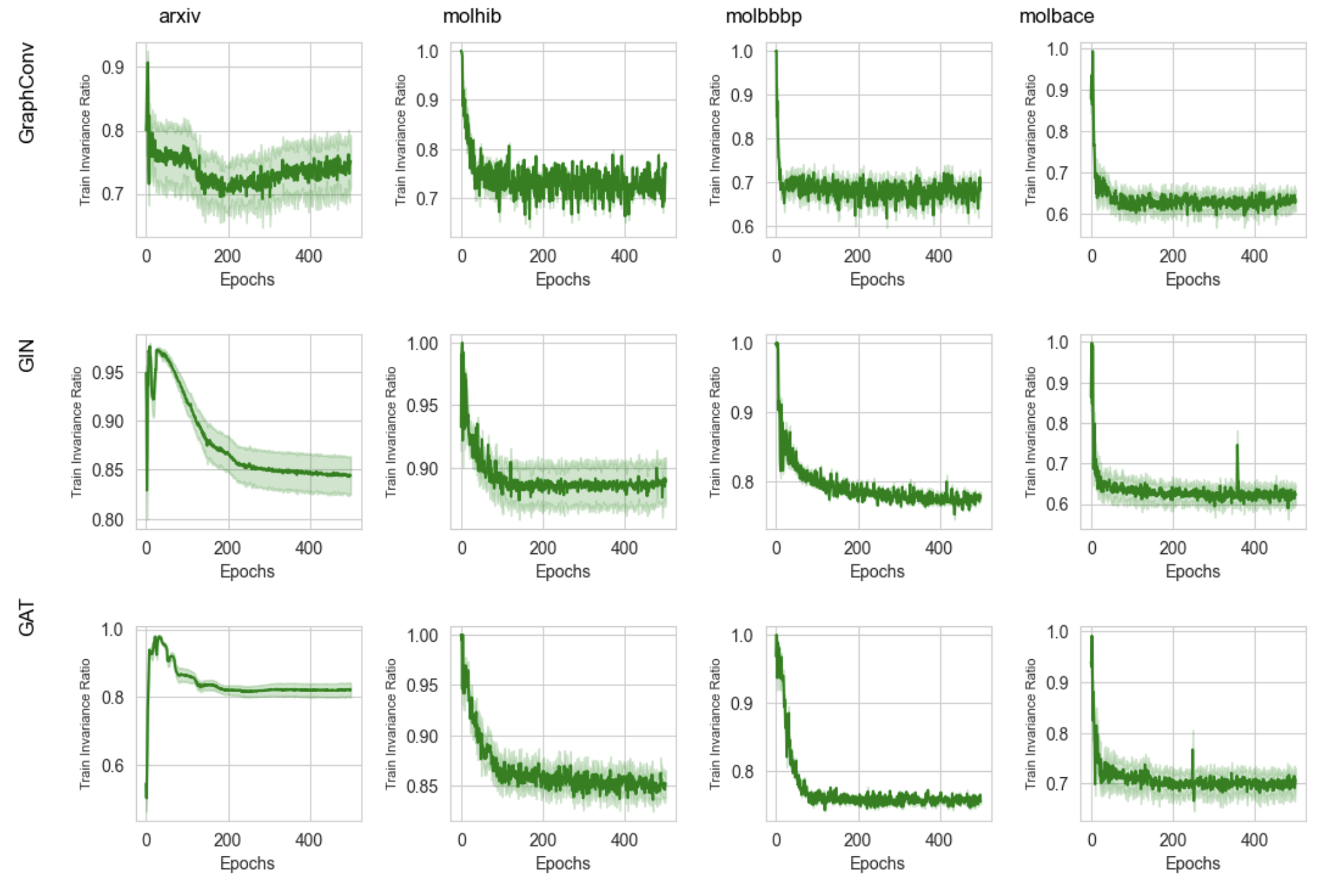} \\

    \caption{Invariance ratios over different GNNs with RNI, on real-world datasets, using 8 random channels as IDs. Despite the ID resampling that is applied in every epoch, invariance is not achieved and, in most cases, does not increase during training.}
    \label{Figure:8_inv}
\end{figure*}

\paragraph{Invariance evolution during training.}
\Cref{Figure:8_inv} depicts the invariance trajectories over epochs for the case of $8$-channel IDs. Similar to the $64$-channel case, we observe that standard RNI-based training initially exhibits partial invariance, which gradually degrades as training progresses. In contrast, \textsc{ICON} maintains high invariance ratios across epochs, often approaching full invariance by convergence. This consistent pattern confirms that \textsc{ICON} successfully constrains the model to remain invariant to the values of random IDs during optimization, independent of architectural choices or ID dimensionality.

\begin{table*}[h!]
\fontsize{7.6}{10}\selectfont
\centering
\setlength{\tabcolsep}{0.3em}
\caption{Performance across OGB datasets for different models with $8$ random channels as IDs. 
We report mean ROC-AUC and standard deviation for the \textsc{molhiv}, \textsc{molbace}, and \textsc{molbbbp} datasets, 
mean accuracy and standard deviation for the \textsc{arxiv} and \textsc{squirrel} datasets, 
and Hits@50 and standard deviation for the \textsc{collab} dataset. 
For each model we report the performance with no IDs, with RNI, and with \ourmethod. 
The best-performing approach within the standard deviation margin is highlighted in bold. 
In all cases, \ourmethod~outperforms RNI and maintains the same trends observed in the main paper.}
\label{tab:wd}
\vskip 0.15in
\begin{tabular}{lcccccc}
\toprule
\textbf{Model} 
& \textsc{ogbn-arxiv} 
& \textsc{ogbg-molhiv} 
& \textsc{ogbg-molbace} 
& \textsc{ogbg-molbbbp} 
& \textsc{ogbl-collab} 
& \textsc{squirrel} \\
\midrule
GraphConv               & 72.05$_{\pm 1.13}$ & \textbf{70.27$_{\pm 2.51}$} & 62.58$_{\pm 1.98}$ & 63.39$_{\pm 0.17}$ & 46.15$_{\pm 1.01}$ & 32.4$_{\pm 1.7}$ \\
GraphConv + RNI         & 70.89$_{\pm 1.42}$ & 55.87$_{\pm 2.33}$ & 51.78$_{\pm 2.14}$ & 58.92$_{\pm 1.91}$ & 40.02$_{\pm 3.42}$ & 28.8$_{\pm 2.0}$ \\
GraphConv + \ourmethod  & \textbf{73.41$_{\pm 1.14}$} & \textbf{70.19$_{\pm 2.01}$} & \textbf{65.19$_{\pm 1.13}$} & \textbf{66.78$_{\pm 1.02}$} & \textbf{47.92$_{\pm 1.68}$} & \textbf{34.7$_{\pm 1.4}$} \\
\midrule
GIN                     & 76.30$_{\pm 1.58}$ & \textbf{72.09$_{\pm 1.01}$} & 73.02$_{\pm 1.27}$ & \textbf{60.86$_{\pm 1.05}$} & 43.87$_{\pm 0.94}$ & 30.2$_{\pm 1.5}$ \\
GIN + RNI               & 71.46$_{\pm 1.88}$ & 70.83$_{\pm 1.92}$ & 62.19$_{\pm 2.39}$ & 53.12$_{\pm 1.78}$ & 39.12$_{\pm 2.25}$ & 26.9$_{\pm 1.8}$ \\
GIN + \ourmethod        & \textbf{79.03$_{\pm 0.87}$} & 71.79$_{\pm 1.33}$ & \textbf{76.52$_{\pm 1.04}$} & 60.21$_{\pm 2.08}$ & \textbf{45.61$_{\pm 1.09}$} & \textbf{33.0$_{\pm 1.1}$} \\
\midrule
GAT                     & \textbf{72.76$_{\pm 1.30}$} & 72.60$_{\pm 2.49}$ & \textbf{71.01$_{\pm 2.83}$} & \textbf{59.42$_{\pm 2.52}$} & 44.23$_{\pm 1.23}$ & 29.8$_{\pm 2.0}$ \\
GAT + RNI               & 62.04$_{\pm 3.04}$ & 54.11$_{\pm 2.04}$ & 63.02$_{\pm 1.48}$ & 58.62$_{\pm 1.93}$ & 40.73$_{\pm 1.96}$ & 27.2$_{\pm 1.5}$ \\
GAT + \ourmethod        & 70.59$_{\pm 2.91}$ & \textbf{76.33$_{\pm 1.20}$} & 68.10$_{\pm 1.42}$ & \textbf{63.14$_{\pm 1.23}$} & \textbf{45.02$_{\pm 1.33}$} & \textbf{32.4$_{\pm 1.2}$} \\
\midrule
TransformerConv          & 73.15$_{\pm 1.39}$ & \textbf{70.11$_{\pm 1.49}$} & 74.91$_{\pm 2.04}$ & \textbf{60.92$_{\pm 1.37}$} & 44.67$_{\pm 1.10}$ & 33.6$_{\pm 1.4}$ \\
TransformerConv + RNI    & 70.34$_{\pm 1.57}$ & 67.21$_{\pm 1.44}$ & 64.78$_{\pm 1.89}$ & 55.01$_{\pm 1.81}$ & 42.02$_{\pm 1.90}$ & 30.8$_{\pm 1.6}$ \\
TransformerConv + \ourmethod & \textbf{75.62$_{\pm 2.02}$} & \textbf{72.98$_{\pm 1.18}$} & \textbf{74.26$_{\pm 1.51}$} & \textbf{59.01$_{\pm 1.79}$} & \textbf{46.21$_{\pm 1.15}$} & \textbf{35.1$_{\pm 1.2}$} \\
\midrule
GraphGPS                 & 74.80$_{\pm 1.99}$ & \textbf{72.75$_{\pm 2.11}$} & 69.94$_{\pm 1.51}$ & \textbf{58.17$_{\pm 2.01}$} & 47.12$_{\pm 0.88}$ & \textbf{37.4$_{\pm 1.2}$} \\
GraphGPS + RNI           & 70.91$_{\pm 2.11}$ & 61.02$_{\pm 3.74}$ & 67.91$_{\pm 1.43}$ & 56.71$_{\pm 2.36}$ & 44.72$_{\pm 1.27}$ & 33.6$_{\pm 1.6}$ \\
GraphGPS + \ourmethod    & \textbf{76.81$_{\pm 2.12}$} & \textbf{73.92$_{\pm 1.68}$} & \textbf{71.84$_{\pm 1.47}$} & \textbf{60.12$_{\pm 1.78}$} & \textbf{48.02$_{\pm 0.96}$} & \textbf{36.5$_{\pm 1.0}$} \\
\bottomrule
\end{tabular}
\vskip 0.15in
\end{table*}

\section{Additional Experimental Details} \label{app:experimental_settings}

\paragraph{Real-world datasets}
The datasets' statistics are summarized in \Cref{tab:ogb_summary}. 
They include both node- and graph-level classification tasks, as well as a link prediction benchmark. 
For the node classification task, we use \textsc{ogbn-arxiv} and the heterophilous \textsc{squirrel} dataset. 
For graph-level prediction, we use the molecular property datasets \textsc{ogbg-molhiv}, \textsc{ogbg-molbace}, and \textsc{ogbg-molbbbp}. 
Finally, \textsc{ogbl-collab} is employed for link prediction. 
All OGB datasets follow the official OGB train/validation/test splits~\citep{NEURIPS2020_fb60d411}. For the \textsc{squirrel} dataset~\citep{10.1093/comnet/cnab014}, we use the standard train/validation/test splits provided by~\citet{Pei2020Geom-GCN}.

\begin{table}[ht]
    \centering
    \caption{Summary Statistics of OGB Datasets}
    \label{tab:ogb_summary}
    \vskip 0.15in
    \begin{tabular}{lccccc} 
    \toprule
    \textbf{Dataset} & \textbf{\# Graphs} & \textbf{Avg \# Nodes} & \textbf{Avg \# Edges} & \textbf{\# Node Features} & \textbf{\# Classes} \\
    \midrule
    \textsc{ogbn-arxiv}   & 1       & 169,343   & 1,166,243  & 128 & 40 \\
    \textsc{ogbg-molhiv}  & 41,127  & 25.5      & 27.5       & 9   & 2  \\
    \textsc{ogbg-molbace} & 1,513   & 34.1      & 36.9       & 9   & 2  \\
    \textsc{ogbg-molbbbp} & 2,039   & 24.1      & 26.0       & 9   & 2  \\
    \textsc{ogbl-collab}  & 1       & 235,868   & 1,285,465  & 128 & 2$^\dagger$ \\
    \textsc{squirrel}     & 1       & 5,201     & 217,073    & 2,089 & 5 \\
    \bottomrule
    \end{tabular}
    \vskip 0.05in
    \smallskip
    \raggedright
    \footnotesize{$^\dagger$ For \textsc{ogbl-collab}, the task is link prediction rather than classification, where edges between authors are predicted based on collaboration features.}
\end{table}

\paragraph{isInTriangle}
The isInTriangle task is a binary node classification where the goal is to determine whether a given node is part of a triangle.
The dataset consists of 100 graphs with 100 nodes each, generated using the preferential attachment (BA) model \citep{badist}, in which graphs are constructed by incrementally adding new nodes with $m$ edges and connecting them to existing nodes with a probability proportional to the degrees of those nodes. We adopt an inductive setting, where the graphs used during testing are not included in the training set. 
We used $m=2$ for all the training graphs.

  \paragraph{EXP and CEXP}
  We followed the protocol of \citet{ijcai2021p291} and used Adam optimizer with a learning rate of $1e-4$. The network has $8$ GraphConv layers followed by a sum-pooling and a $3$ layer readout function. All hidden layers have $64$ dimensions, and we used $64$ random features and we did not discard the original features in the data. 

\paragraph{Hyper-Parameters}
For all experiments, we use a fixed dropout rate of 0.1 and ReLU activations.
We tuned the learning rate in $\{10^{-3}, 5\cdot 10^{-4}\}$, batch size in $\{32, 64\}$ (for graph-level tasks), number of layers in $\{3, 5\}$, hidden dimensions in $\{32, 64\}$, and weight decay in $\{10^{-4}, 10^{-5}\}$. For node-level and link prediction tasks (\textsc{ogbn-arxiv}, \textsc{squirrel}, \textsc{ogbl-collab}), we use full-graph training. 
Random seeds were fixed for reproducibility, and results are reported across three independent runs. All models were trained on a single NVIDIA V100 32GB GPU. 

\section{Time Complexity Analysis}
\label{app:complexity}

In this section, we analyze the computational complexity introduced by \textsc{ICON} relative to a standard GNN trained with RNI. Recall that \textsc{ICON} adds an explicit regularization term encouraging invariance to random ID values by comparing embeddings produced under two independent ID assignments.

Let the base GNN-R (with random IDs) consist of $L$ message-passing layers with hidden dimensions $\{d_\ell\}_{\ell=1}^L$. Denote by $|V|$ and $|E|$ the number of nodes and edges in the input graph, respectively. The computational complexity of a single forward pass of a message-passing GNN is
\[
O\!\left(\sum_{\ell=1}^L |E|d_{\ell-1} + |V|d_\ell \right),
\]
which corresponds to the cost of aggregating and updating node representations across all layers.

\paragraph{Additional cost of \textsc{ICON}.}
\textsc{ICON} introduces one additional forward pass per training batch to evaluate the same network under a second random ID assignment $I'$. Let $T_f$ denote the cost of one forward pass and $T_b$ the cost of one backward pass. The total training cost per iteration can then be expressed as:
\[
T_{\text{RNI}} = T_f + T_b, \qquad
T_{\text{ICON}} = 2T_f + 2T_b + O(|V|d_L),
\]
where the additional $O(|V|d_L)$ term arises from computing the $\ell_2$ regularization term
\(\|H^{(L)}_1 - H^{(L)}_2\|_2^2\) at the final layer. The backward pass cost doubles because the regularization term depends on embeddings from both forward passes, requiring gradient computation through both computation graphs.

Therefore, the overall time complexity of \textsc{ICON} remains in the same asymptotic order as the base GNN, that is,
\[
O(|E|d),
\]
but with a multiplicative constant of approximately $2\text{--}3\times$ due to the extra forward and backward passes. The precise overhead depends on implementation details and hardware-level parallelization efficiency. In practice, the cost can be mitigated because both forward passes share the same graph structure and model parameters, allowing for partial reuse of message computations and efficient batched evaluation on GPUs.

Overall, \textsc{ICON} preserves the asymptotic time complexity of standard message-passing GNNs, with only a modest constant factor overhead. Specifically:
\begin{itemize}
    \item \textbf{Asymptotic complexity:} $O(|E|d)$ per forward pass, same as standard GNNs.
    \item \textbf{Constant overhead:} one extra forward pass, one extra backward pass, and a lightweight vector difference at the final layer.
    \item \textbf{Empirical runtime:} In our experiments, \textsc{ICON} increased the total wall-clock training time by approximately $2\text{--}3\times$, consistent with the doubling of both forward and backward computations, while often converging in fewer epochs due to improved invariance and optimization stability.
\end{itemize}

Overall, \textsc{ICON} provides a favorable trade-off between a small computational overhead and substantial gains in ID-invariance, generalization, and convergence speed.

\section{Future work}
\label{app:future}
An intriguing open question is whether two layers of a GNN with unique identifiers (IDs) suffice to achieve additional expressiveness beyond the 1-WL test.
We showed that using ID-invariance on all layers does not enhance expressiveness, whereas enforcing invariance solely in the final layer does. Fully understanding the conditions on layers regarding ID-invariance and identifying which functions can be computed under various settings presents an interesting direction for future research.
Our experiments show that ID utilization can be improved with {\ourmethod}, by \textit{learning} to produce an invariant network. Another interesting open question is how to design GNNs that benefit from the added theoretical expressiveness of IDs \textit{and} are invariant to IDs \textit{by design}.

\end{document}